\documentclass[lettersize,journal]{IEEEtran}
\usepackage{amsmath,amsfonts}
\usepackage{array}
\usepackage[caption=false,font=footnotesize,labelfont=rm,textfont=rm]{subfig}
\usepackage{textcomp}
\usepackage{stfloats}
\usepackage{url}
\usepackage{verbatim}
\usepackage{graphicx}
\usepackage{cite}
\newcommand{\Ours}{{Light-SLAM}}
\usepackage{multirow} 
\usepackage{threeparttable} 
\usepackage{booktabs} 
\usepackage{algorithm}  
\usepackage{algpseudocode}  

\def\etal{{\em et al.}} 

\def\etal{{\em et al.}}

\begin{document}

\title{\Ours: A Robust Deep-Learning Visual SLAM System Based on LightGlue under Challenging Lighting Conditions
}

\author{Zhiqi~Zhao,~Chang~Wu,~\IEEEmembership{Senior Member,~IEEE},~Xiaotong~Kong,~Zejie~Lv,~Xiaoqi~Du~and~Qiyan~Li
\thanks{Zhiqi Zhao, Chang Wu, Xiaotong Kong, Xiaoqi Du and Qiyan Li are with the Institute of Information and Communication Engineering, University of Electronic Science and Technology of China (UESTC), Chengdu 611731, China (email:202252012118@std.uestc.edu.cn; changwu@uestc.edu.cn; m13086663208@163.com; 202222010904@std.uestc.edu.cn; llittle7@163.com; qyan.lieestd@gmail.com).

Zhiqi Zhao and Chang Wu have the same contribution to this paper as co-first authors, and Chang Wu is the corresponding author.
}
}




\maketitle

\begin{abstract}
Simultaneous Localization and Mapping (SLAM) has become a critical technology for intelligent transportation systems and autonomous robots and is widely used in autonomous driving. However, traditional manual feature-based methods in challenging lighting environments make it difficult to ensure robustness and accuracy. Some deep learning-based methods show potential but still have significant drawbacks. To address this problem, we propose a novel hybrid system for visual SLAM based on the LightGlue deep learning network. It uses deep local feature descriptors to replace traditional hand-crafted features and a more efficient and accurate deep network to achieve fast and precise feature matching. Thus, we use the robustness of deep learning to improve the whole system. We have combined traditional geometry-based approaches to introduce a complete visual SLAM system for monocular, binocular, and RGB-D sensors. We thoroughly tested the proposed system on four public datasets: KITTI, EuRoC, TUM, and 4Season, as well as on actual campus scenes. The experimental results show that the proposed method exhibits better accuracy and robustness in adapting to low-light and strongly light-varying environments than traditional manual features and deep learning-based methods. It can also run on GPU in real time.
\end{abstract}


\section{Introduction}
\IEEEPARstart{S}{imultaneous} Localization and Mapping (SLAM) has become a popular area of research in intelligent transportation systems and autonomous robots over the last two decades. It solves the problem of robots using sensors to acquire information about their environment, allowing them to localize themselves in unknown environments and map their surroundings simultaneously
\cite{tsintotas2022revisiting,lee2016ground}. Real-world applications such as mobile robots, autonomous driving, and VR devices widely use SLAM. Due to the low cost of cameras and their ability to capture rich environmental information, researchers have extensively studied Visual SLAM (VSLAM).

In a VSLAM system, the front-end visual odometry (VO) system determines the robot's current position and orientation by processing camera image information and acquiring motion relationships as the camera moves. Hand-crafted features are the dominant method used in the VO \cite{mur2015orb,mur2017orb,davison2007monoslam,klein2009parallel}. By extracting features from multiple images and matching their descriptors, we can obtain camera pose estimation information for the VO. However, when confronted with some challenging scenes (e.g., low-light, dramatically changing light, and low-texture scenes), traditional methods based on manual features may not be able to recognize and correlate feature information in the current surroundings.

With the development of deep learning, using deep neural networks (DNN) for camera motion estimation has become increasingly popular
\cite{cui2021deep,feng2020deep}. These networks are trained on large datasets and used to build an end-to-end system, replacing traditional VO) or entire VSLAM system \cite{li2018undeepvo,wang2017deepvo,teed2021droid}. In addition, some research works have attempted to enhance the conventional VSLAM system by using DNN to replace some modules of the traditional process of VO. Using DNN to extract features to replace classic hand-designed features to improve system robustness has become a dominant trend in related research 
\cite{kang2019df,tang2019gcnv2,deng2019comparative}.

In specific scenes, these approaches demonstrate the excellent potential of deep learning in the field of SLAM. However, deep learning and SLAM fusion methods must be more mature and have apparent shortcomings. When faced with scenes in unknown environments, most deep learning methods perform poorly due to their heavy reliance on the specific dataset used for training. Furthermore, most proposed methods must comprehensively evaluate the system's robustness under different scenes, including challenging ones. More importantly, there are also severe challenges regarding computational efficiency when taking advantage of deep learning techniques. Such a design severely limits the system's performance and often fails to meet SLAM systems' stability and real-time requirements.

Thus, this paper proposes a novel VSLAM system that utilizes deep local feature descriptors to replace traditional manual features. The system uses the LightGlue network
\cite{lindenberger2023lightglue} for feature tracking, significantly improving localization accuracy while operating in real-time. Experiments show that even in challenging low-light scenes, our proposed method still works well in overcoming the difficulty that popular algorithms such as ORB-SLAM2 may not work correctly to track the camera position and construct environment maps accurately.

The main contributions are as follows: 
\begin{enumerate}
    \item a proposed hybrid approach for visual SLAM based on the LightGlue deep learning network: \Ours. It utilizes deep local feature descriptors instead of traditional hand-crafted features and achieves precise feature matching quickly through a more accurate and efficient deep network. This method enhances the system's robustness in adapting to challenging environments such as low-light conditions.

    \item proposes an optimization algorithm for feature extraction using the image pyramid model. It assigns the feature extraction tasks of each image layer to different processing units in parallel, aiming to improve the efficiency of image feature tracking and the real-time performance of the system.
    
    \item proposes a precise matching method for stereo image frames using deep neural networks to replace traditional coarse matching methods for binocular stereo. This method aims to achieve more accurate stereo-visual reconstruction and depth estimation by improving the accuracy of data association between image pixel points and spatial map points. 

    \item the \Ours~method is tested and evaluated on four types of publicly available datasets: KITTI 
    \cite{geiger2013vision}, EuRoC 
    \cite{burri2016euroc}, TUM 
    \cite{sturm2012benchmark} and 4Season 
    \cite{wenzel20214seasons}, as well as real campus scene sequences. Experimental results show that \Ours~is superior in adapting to the environment with different illumination variations at different times of day, significantly improving the robustness and accuracy of the system while being able to run in real-time on GPU.
\end{enumerate}

In the rest of this paper, we discuss related work in Section II. We present the architecture of \Ours~system and the primary methodology in Section III. We then offer the evaluation results of \Ours~on different datasets in Section IV. Finally, we summarize the findings and future research directions in Section V.

\section{Related Work}

\subsection{Feature-based Traditional Methods}

The feature-based method is one of the most commonly used methods in vision SLAM. The main advantage of the proposed method is its runtime stability, making it a relatively mature solution in the current visual SLAM field. However, feature extraction may fail in environments without texture, low illumination, or drastic lighting changes.

The feature extraction stage includes a variety of traditional algorithms, including Harris corner point 
\cite{harris1988combined}, SIFT 
\cite{lowe2004distinctive}, SURF 
\cite{bay2006surf}, FAST 
\cite{rosten2008faster}, ORB 
\cite{rublee2011orb}, and so on. Davison \etal
\cite{davison2007monoslam} proposed the first real-time monocular visual SLAM system, MonoSLAM, based on features. This method lacks global optimization and closed-loop detection, and the computational cost increases proportionally with the size of the environment, which has the drawbacks of smaller application scenes and easy loss of feature points. Klein \etal 
\cite{klein2009parallel} proposed Parallel Tracking Mapping (PTAM), where the tracking thread extracts FAST corner points, estimates camera motion in real-time based on the image data, and performs the mapping thread in parallel for the prediction of 3D feature point positions. Although it performs better than Mono-SLAM, it is still limited to small-scale scenes, and motion blur easily causes loss of tracking in the tracking phase. Labbe \etal 
\cite{labbe2019rtab} proposed RTAB- MAP, which aims to provide a method with real-time constraints, proposes a loopback detection module with the function of processing memory space, and can handle long and large-scale online mapping operations. Fang \etal 
\cite{fang2018he} proposed a HE-SLAM method for light changes to extract effective ORB feature points in low-contrast images. However, the manually designed feature information is still difficult to extract when the illumination changes drastically. ORB-SLAM 
\cite{mur2015orb} and ORB-SLAM2 
\cite{mur2017orb} proposed by Mur-Artal \etal are complete visual SLAM schemes based on the feature-point method. They achieve real-time localization and map building in large indoor and outdoor environments by extracting ORB features and computing the corresponding descriptors and feature matching using Hamming distance, supporting re-localization and auto-initialization. The previous version only supports monocular camera operation, which suffers from scale drift and fails when performing pure rotation. They then proposed the ORB-SLAM2 method applied to stereo and RGB-D cameras.

However, their performance is unstable and even tracking loss is possible in the face of complex and changing scenes, such as low light, drastic light changes, and insufficient texture.

\subsection{Deep Learning Enhanced VSLAM Methods}

In recent years, deep learning has become widely popular in the computer vision community, and there is a significant trend in the research of visual SLAM systems based on deep learning. Deep learning provides powerful tools that can integrate into many modules of vision SLAM to cope with working in challenging and complex environments.

Wang \etal 
\cite{wang2017deepvo} proposed a novel end-to-end monocular VO framework using deep recurrent convolutional neural networks (RCNN) called DeepVO, the first end-to-end approach for monocular VO based on DNN. Their approach uses deeply learned geometric feature representations combined with sequential modeling using convolutional neural networks (CNN) and recurrent neural networks (RNNs) to estimate poses directly from raw RGB images. DeepVO can combine geometric shapes with learned knowledge models without a priori information or parameters to recover absolute proportions and be used to augment the VO. However, it cannot yet be used to replace traditional geometric feature-based VO methods. Li \etal 
\cite{li2018undeepvo} proposed another monocular end-to-end method based on unsupervised learning called UnDeepVO. Unlike DeepVO proposed by Wang \etal, this method does not require ground truth as it operates in an unsupervised manner. Parisotto \etal 
\cite{parisotto2018global} proposed an end-to-end system with an architecture similar to DeepVO, using a Neural Graph Optimization (NGO) step containing an Attention Machine phase instead of a Long Short-Term Memory (LSTM) layer for closed-loop-like drift correction. They outperform the DeepVO approach in simulated 2D and 3D maze environments but still lack test validation in more public dataset scenes.

The problem with some traditional feature-based visual SLAM methods is that they cannot effectively adapt to various complex and unpredictable scenes. Therefore, many researchers have used DNN to extract deep features at different stages of an image, replacing traditional hand-crafted features to achieve higher recognition and matching rates to solve the problem.

\begin{figure*}[t]
\centering
\includegraphics[width=0.8\linewidth]{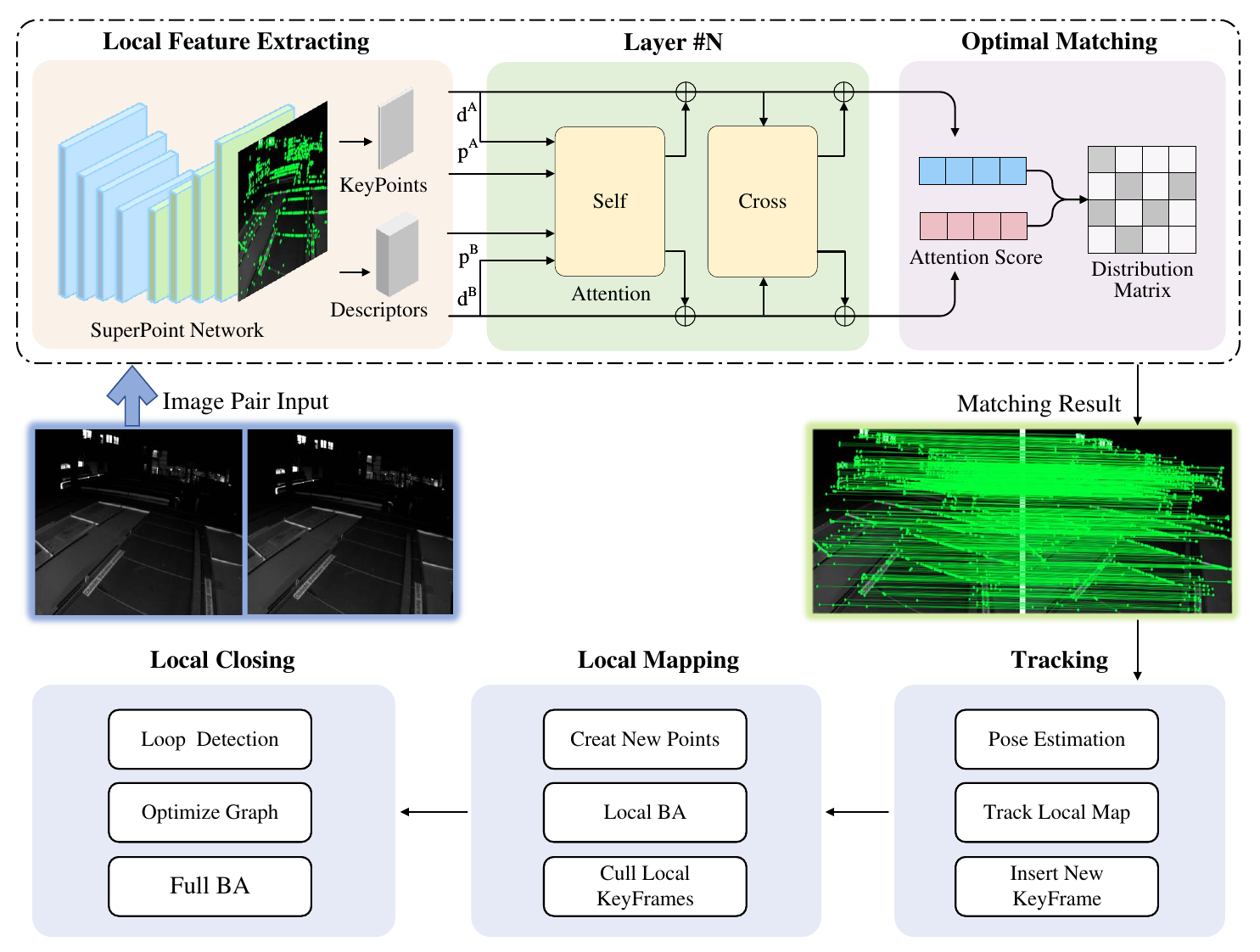}
\caption{The \Ours~system framework.We extract local feature points and compute corresponding descriptors using a SuperPoint network, denoted as $(d, p)$ for an image pair input system. We use a multi-layer network based on self-attention and cross-attention to update the feature state. We characterize the similarity and matchability between the points based on the attention score and complete the assignment between the predicted points to obtain the optimal matching result. Inputs to the threads running parallel in the system: tracking, local mapping, and loop closing.}
\label{fig_1_framework}
\end{figure*}

In this area, Tateno \etal 
\cite{tateno2017cnn} proposed a CNN-based approach called CNN-SLAM. The method's main idea is to segment the image captured by the camera into smaller parts for a better understanding of the environment. It performs depth prediction at the location of the image where monocular SLAM methods tend to fail, e.g., along low-texture regions, thus overcoming one of the main limitations of monocular SLAM. Bruno \etal 
\cite{bruno2021lift} proposed LIFT-SLAM, which extends the pipeline of the ORB-SLAM system and uses CNN to extract features from images, using the learned features to provide denser and more accurate matching. However, this impacts real-time performance due to its computationally intensive pipeline and unoptimized CNN design. Stephen \etal 
\cite{hausler2021patch} proposed Patch-NetVLAD to aggregate and match deep-learned local features defined on a feature space grid through deep learning, which performs well in scenes with varying lighting. Tang \etal 
\cite{tang2018geometric} proposed a GCN neural network algorithm based on deep learning. Extracting feature points and descriptors end-to-end and using deep metric learning to achieve feature association improves the efficiency of image feature extraction, with the disadvantage that the feature extraction failure problem still occurs in the face of scenes with drastic lighting changes. Subsequently, they also proposed another version \etal 
\cite{tang2019gcnv2}, which combines CNN and RNN to train geometric coping networks, inspired by SuperPoint
\cite{detone2018superpoint} while detecting key points to generate their corresponding descriptors, with higher and more stable feature extraction accuracy. However, the algorithm has poor real-time performance and more mismatches during feature matching.

In summary, existing feature-based traditional visual SLAM methods have difficulty adapting to low-light and dramatically changing lighting environments. Deep learning-based methods show some potential in such scenes but still have obvious drawbacks. Therefore, this paper proposes a hybrid approach to visual SLAM based on the LightGlue deep learning network. We use deep local feature descriptors to replace the traditional hand-crafted features, and we use deep learning to improve the robustness of the whole SLAM system.

\section{\Ours}

\subsection{System Framework}

The framework of \Ours~system is shown in Fig. \ref{fig_1_framework}; similar to the ORB-SLAM2 framework, the system mainly contains three threads running in parallel: tracking, local mapping, and loop closing.

In the tracking thread, for each frame of image input to the system, we use a deep learning network to extract local feature descriptors to replace the traditional hand-designed features. We then achieve fast and accurate matching with a network based on an attention mechanism. Then, a constant velocity motion model estimates the initial camera pose. At the same time, we track the local map and perform a reprojection search to match the local map points. Finally, new keyframes are inserted into the map or not according to the thread state judgment.

In the local mapping thread, when new keyframes are received, new map points are first created by triangulating the features of the associated keyframes in the common view. Then, we perform local BA to optimize the camera position. As the number of keyframes increases, we reduce the system's complexity by eliminating redundant local keyframes.

In the loop closing thread, we first detect the last keyframe processed in the previous thread to satisfy the loopback condition by acquiring it. If loopback is detected, we immediately compute the similarity transform of the current keyframe to the loop closure frame to obtain the accumulated error in the loopback and optimize the bitmap to achieve global consistency. Finally, we incorporate the full BA optimization to achieve the optimal solution.

\subsection{Deep Local Feature Descriptor Extraction and Matching}

\begin{figure}[t]
\centering
\includegraphics[width=\linewidth]{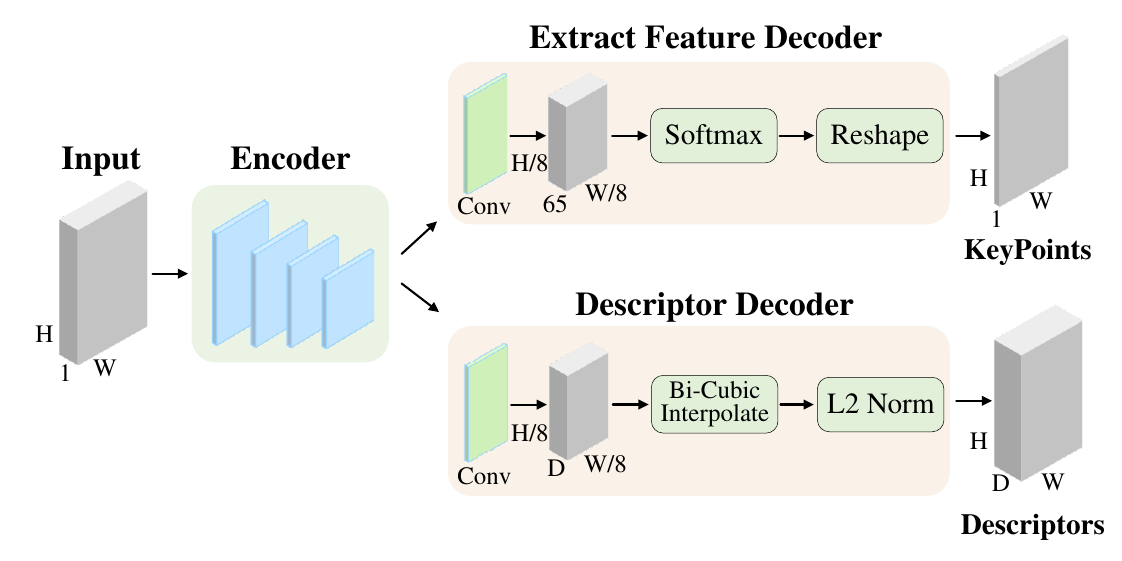}
\caption{The deep learning feature extractor architecture. It consists of a shared encoder and two decoders. Both decoders operate on a shared space-reduced representation of the input. One trains to detect feature points, and the other to detect corresponding descriptors.}
\label{fig_2_superpoint_framework}
\end{figure}

In the feature extraction phase, we use a fully convolutional neural network architecture that performs a single forward pass over a full-size image to extract SuperPoint feature points with fixed-length descriptors. As shown in Fig. \ref{fig_2_superpoint_framework}.

First, for an image of size W × H, the feature detector computes W/8 × H/8 × 65 and outputs a tensor of size W × H. The 65 channels contain a localized, non-overlapping 8 × 8 pixel grid region and an additional 'non-feature' bin. We remove this dustbin dimension using channel softmax and apply a shaping function to output a tensor of size W × H. The descriptor decoder uses a semi-dense rather than dense design to reduce training parameters and maintain run-time processability. It then performs bicubic interpolation of the descriptors and L2 normalization to output the descriptors finally.

SuperPoint demonstrates its robustness when adapting to different challenging scenarios. Replacing traditional hand-crafted features with this deep feature can achieve higher recognition and matching rates, overcoming the poor performance of some traditional feature-based visual SLAM methods in challenging scenes. More importantly, SuperPoint is robust in adapting to various complex scenes.

We associate each local feature $i$ in an image $I\in\{A,B\}$ with a state $\mathbf{x}_i^I \in\mathbb{R}^d$. The state is initialized with the corresponding visual descriptor $\mathbf{x}_i^I \leftarrow \mathbf{d}_i^I$ and subsequently updated by each layer. We define a layer as a sequence of self-attention and cross-attention units. In each attention unit, the multilayer perceptron (MLP) computes all feature points in both images in parallel, updating the feature states based on the messages $\mathbf{m}_i^{I \leftarrow S}$ aggregated from the source image $S \in\{A, B\}$.

Then, the attention score is computed, and predicted assignments are made based on the updated status of any layer. The similarity matrix $\mathbf{S} \in \mathbb{R}^{M \times N}$ between the feature points of the two images is first computed:
\begin{equation}
\label{deqn_ex5}
\mathbf{S}_{i j}=\operatorname{Linear}\left(\mathbf{x}_i^A\right)^{\top} \text { Linear }\left(\mathbf{x}_j^B\right) \ \forall(i, j) \in \mathcal{A} \times \mathcal{B}
\end{equation} where $\operatorname{Linear}(\cdot)$ is a learned linear transformation with bias. 
Then, we calculate the likelihood of having a corresponding point for each feature point $i$, the matching score 
\begin{equation}
\label{deqn_ex6}
\sigma_i=\operatorname{Sigmoid}\left(\operatorname{Linear}\left(\mathbf{x}_i\right)\right) \in[0,1]
\end{equation}

If the match fails, then $\sigma_i \rightarrow 0$.

We combine the similarity and match scores in an assignment matrix $\mathbf{P}$ as
\begin{equation}
\label{deqn_ex7}
\mathbf{P}_{i j}=\sigma_i^A \sigma_j^B \operatorname{Softmax}\left(\mathbf{S}_{k j}\right)_i \operatorname{Softmax}\left(\mathbf{S}_{i k}\right)_j
\end{equation}

When we predict a pair of points $(i, j)$ to be matchable, and their similarity is higher than any other point in the two images, we optimally match the two points.

\subsection{Optimized Parallel Image Pyramid Model}
To improve the robustness and efficiency of the system tracking, we adopt the popular SLAM system to achieve scale invariance by constructing an image pyramid. We perform multi-scale feature point extraction and matching to enhance the algorithm's robustness to object scale changes. However, these additional operations incur a sizeable computational overhead, which affects the system's efficiency, especially when dealing with large-size images.

Therefore, we propose optimizing the algorithm for feature extraction on the image pyramid model. The feature extraction of multiple layers of images is accomplished in parallel to improve the overall processing speed and the real-time performance of the system. The computational resources are fully utilized by assigning the extraction tasks of each layer of images to different processing units, avoiding the bottleneck of serial processing, thus effectively speeding up the extraction of feature points, improving the efficiency of system tracking, and solving many problems that are difficult to meet the real-time requirements of SLAM systems based on deep learning methods.

\begin{figure}[!t]
\centering
\includegraphics[width=0.8\linewidth]{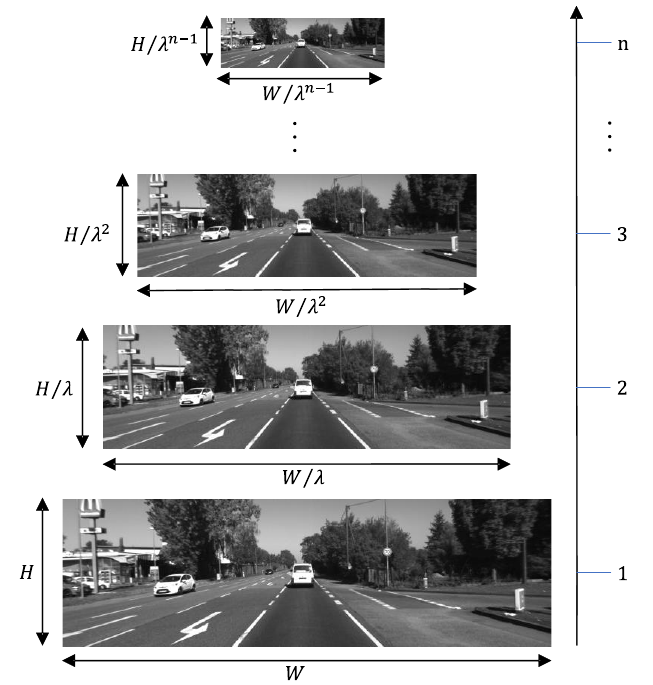}
\caption{Optimized parallel image pyramid model. The input image is progressively down-sampled with a scaling factor $\lambda$ to obtain images of different resolutions, and feature point extraction is done in parallel at each layer of the pyramid. }
\label{fig_3_image_pyramid}
\end{figure}

We show the pyramid construction process in Fig. \ref{fig_3_image_pyramid}. From the bottom layer to the top layer is a downsampling process; the image of the bottom layer is the original image inputted by the system, the width is $W$, the height is $H$, and its resolution is $W \times H$. The scaling factor of the image pyramid is $\lambda(0<1 / \lambda<1)$, and the number of layers is $n$. We set $\lambda$ to 1.2 and $n$ to 8 . We then decrease the width and height of each layer in order by the power of $\lambda$ until we reach $n$ layers, completing the construction of the pyramid.

\begin{figure}[!t]
\centering
\includegraphics[width=\linewidth]{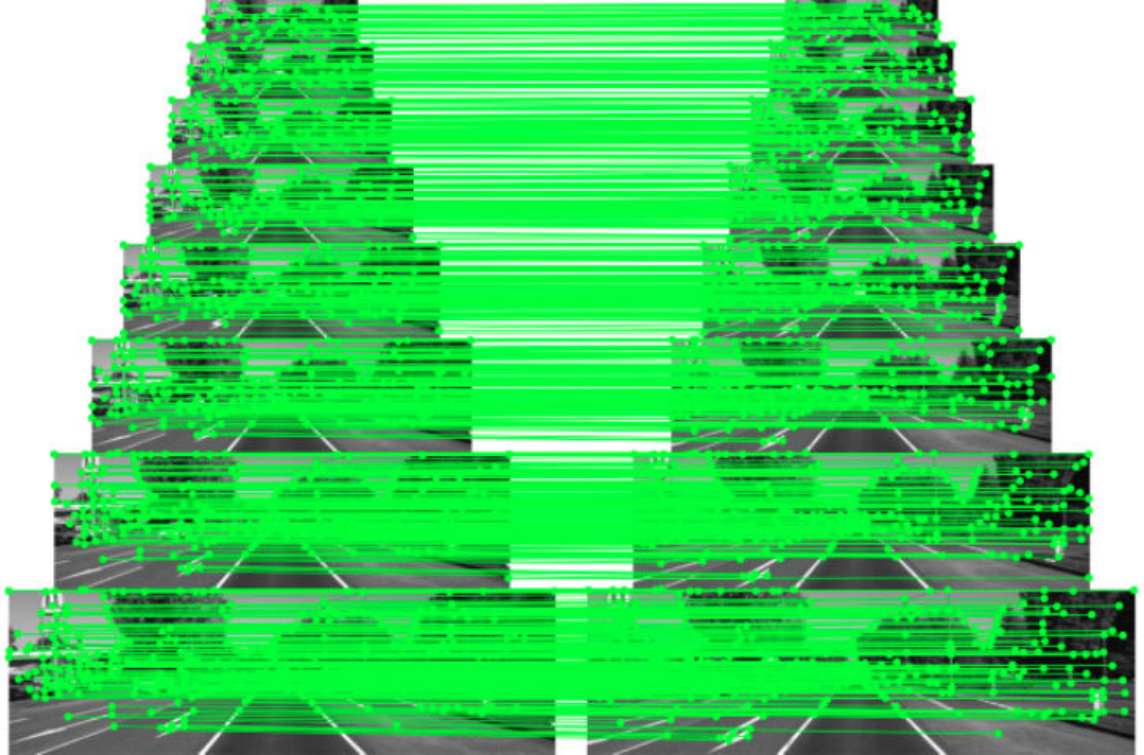}
\caption{Multi-scale feature point matching results. Correlate the number of feature points that should be extracted for each layer of the image with the size of the area of that layer, obtaining feature points that cover a wider range of scales.}
\label{fig_4_match_pyramid}
\end{figure}

Fig. \ref{fig_4_match_pyramid} shows the proposed image pyramid for feature extraction and matching. The higher the number of layers in the image pyramid, the lower the resolution of the image and the lower the number of feature points that can be extracted. Therefore, we can relate the number of feature points extracted for each layer to the size of the image area in that layer to generate feature points that cover more scales. We denote the total area of the entire image pyramid as $S$, with:
\begin{equation}
\label{deqn_ex8}
\begin{aligned}
S=&W \times H \times [1 / \lambda]^0+W \times H \times[1 / \lambda]^1 \\
&+\cdots+W \times H \times[1 / \lambda]^{(n-1)} \\
=&\underbrace{W \times H}_C \times \frac{1-[1 / \lambda]^n}{1-[1 / \lambda]}=C \frac{1-[1 / \lambda]^n}{1-[1 / \lambda]}
\end{aligned}
\end{equation}

Then, the number of feature points $N_{a v g}$ to be extracted per unit area is
\begin{equation}
\label{deqn_ex9}
N_{a v g}=\frac{N}{S}=\frac{N}{C\frac{1-[1/\lambda]^{n}}{1-[1/\lambda]}}=\frac{N\{1-[1/\lambda]\}}{C\{1-[1/\lambda]^{n}]}
\end{equation} where $N$ is the total number of feature points extracted from the image pyramid. Then, we can calculate the number of feature points to be extracted for the $\alpha$ layer of the image pyramid as
\begin{equation}
\label{deqn_ex10}
N_{\alpha}=\frac{N\{1-[1/\lambda]\}}{\{1-[1/\lambda]^{n}\}}[1/\lambda]^\alpha
\end{equation}

Specifically, the feature extraction and matching process is shown in Algorithm \ref{alg:pyramid}. Among them, image pyramids of different levels contain similar but slightly different image information, which means that some feature points may be repeatedly detected in pyramids of different levels, thus introducing redundant feature points. Therefore, we perform non-great value suppression culling to avoid the output of duplicate feature points.

\begin{algorithm}
  \caption{Optimized Parallel Image Pyramid Model Algorithm}
  \label{alg:pyramid}
  \begin{algorithmic}[1]
    \Require
        Two consecutive images $I_{prev}$ and $I_{curr}$ with resolution $W \times H$;
    \Ensure 
        Matching feature point pairs $M$.
    \State Construct the image pyramid $G_p$ by downsampling from the bottom to the top with the scaling factor $\lambda$;
    \State Calculate the total area of the entire image pyramid $S$ using formula \ref{deqn_ex8};
    \State Calculate the number of feature points to be extracted per unit area $N_{a v g}$ using formula \ref{deqn_ex9};
     \For{each layer $\alpha$ in $G_p$}
        \State Calculate the number of feature points assigned to the 

            $\alpha$ layer $N_{\alpha}$ using formula \ref{deqn_ex10};
        \State Extract feature points in parallel on $G_{p}$;
        \State Obtain benchmark feature points $p_{i}\in{P}$ and candidate 
        
        repeat feature points $q_{j}\in{Q}$;
    \EndFor
    \For{each point $q_{j}$ in ${Q}$}
        \State Calculate the coordinate distance between $q_{j}$ and $p_{i}$ 
        
        as $dist$;
      \If {$dist$ \textgreater $l$ (where $l$ is the threshold)} 
            \State Calculate the response threshold $\varepsilon_{j}^{\prime}$ for $q_{j}$;
            \State Calculate the local maxima of $\varepsilon_{j}^{\prime}$ as the solution; 
            \State Add the match to the matching pairs as $M$; 
        \Else
            \State Remove $q_{j}$ as a repeat point;
        \EndIf  
            
  	\EndFor
  \end{algorithmic}  
\end{algorithm}

Using the bottom layer of the image pyramid as the base, we select feature points $p_{i}\in\mathbf{P},\mathbf{P}=\{p_{1},p_{2},\ldots,p_{i}\}$ on each pyramid layer as existing feature points. Then calculate the distance between the candidate feature points (including duplicate feature points) $q_{j}\in\mathbf{Q},\mathbf{Q}=\{q_{1},q_{2},\ldots,q_{j}\}$ and the coordinates of $p_{i}$ within a certain grid region where they are located. Then filter the candidate feature points $q_{j}^{\prime}=\{x_{j},y_{j},\varepsilon_{j}\}$ that are smaller than the critical threshold $l$, $\varepsilon_{j}$ is the response threshold. We sort the feature points according to the size of $\varepsilon_{j}$, and according to the non-maximal value suppression algorithm, the local maxima $\varepsilon_{j}^{\prime}$ are searched to get the neighboring feature points at the best position where the feature points are located. The neighboring candidate feature points with low thresholds are eliminated to avoid the output of duplicate feature points.

\subsection{Stereo Depth Estimation Module}

\begin{figure}[h]
\centering
\includegraphics[width=0.95\linewidth]{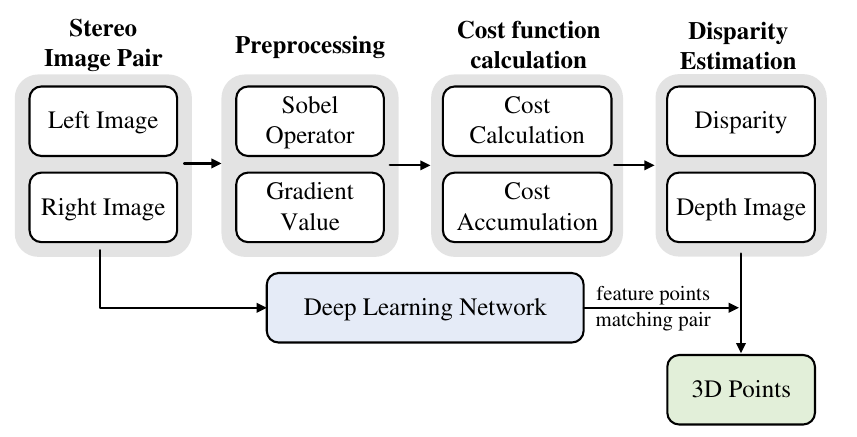}
\caption{The framework for stereo depth estimation module. By preprocessing the input stereo image pairs, calculating the cost function, and obtaining the disparity estimation and the depth image. The accurate matching results of the stereo images using the LightGlue network are compared with the depth image to output the spatial points corresponding to the feature points.}
\label{fig_5_depth_estimate}
\end{figure}

The framework of our proposed stereo depth estimation module is shown in Fig. \ref{fig_5_depth_estimate} . The system first preprocesses the input stereo image pairs and calculates the gradient magnitude at pixel points. Then, the cumulative function of the matching cost is estimated to obtain the best matching surrogate value at each pixel location, and the parallax prediction value is obtained and transformed into a depth map. We utilize the deep learning network to match the stereo image accurately, compare the results with the depth map, and output to get the spatial points corresponding to the feature points. The accuracy of data correlation between image feature points and spatial map points is improved to achieve more accurate stereo-visual reconstruction and depth estimation.

Assuming that the gray value of the left image is $I_{l}(x,y)$, then the gradient magnitude in the horizontal and vertical directions is calculated as:
\begin{equation}
\label{deqn_ex11}
G_{x} =\sum_{i=-1}^{1}\sum_{j=-1}^{1}I_{l}(x+i,y+j)\times S_{x}(i,j)
\end{equation}
\begin{equation}
\label{deqn_ex12}
G_{y} =\sum_{i=-1}^{1}\sum_{j=-1}^{1}I_{l}(x+i,y+j)\times S_{y}(i,j) 
\end{equation} Where $S_x$ and $S_y$ are the horizontal and vertical convolution kernels, respectively. Then, the magnitude of the grey value of the pixel point of the image is:
\begin{equation}
\label{deqn_ex13}
G_{0}=|G_{x}|+|G_{y}|
\end{equation}

For each pixel point $(x,y)$, the coordinates of the pixel corresponding to it in the right image are $(x+d,y)$ , where $d$ is the parallax value. We can define the cost function $C(x,d)$ between the two as:
\begin{equation}
\label{deqn_ex15}
C(x,d)=\sum_{i=-n}^{n}\sum_{j=-n}^{n}|G_{l}(x+i,y+j)-G_{r}(x+d+i,y+j)|
\end{equation}

To reflect the correlation between pixels more accurately, we obtain the best matching surrogate value for each pixel position by accumulating the matching costs along each direction (horizontal, vertical, and diagonal) to select the global minimum cost path. Then, the expression of the cost function is
\begin{equation}
\label{deqn_ex16}
\begin{aligned}
{C}_{r}(x,d) = & {C}(x,d)-\min\limits_{i=d_{\text{min}},\ldots,d_{\text{max}}}C_{r}(x-r,i)\\
&+\min \begin{Bmatrix} C_{r}(x-r,d) \\
C_{r}(x-r,d+1)+P_{1} \\
\min\limits_{i=d_{\text{min}},\ldots,d_{\text{max}}}C_{r}(x-r,i)+P_{2}
\end{Bmatrix} \\
\end{aligned}
\end{equation} Where $P_1$ and $P_2$ are smoothing penalty terms, the second term of the equation is used to eliminate the effects caused by the different path lengths in each direction. The total matching cost is obtained by summing the matching costs in all $r$ directions and obtaining the location of the smallest matching cost, which in turn yields the parallax value of the current pixel point $d$:
\begin{equation}
\label{deqn_ex17}
s(x,d)=\sum_{r}L_{r}(x,d)
\end{equation}

\begin{equation}
\label{deqn_ex18}
d(x,y)=\arg\min_{d}s(x,d)
\end{equation} Where $\underset{d}{\arg\min}$ denotes the $d$ when finding the value that makes $s(x,d)$ take the minimum value, and according to the geometric relationship of binocular vision, the depth the conversion formula of parallax can obtain value to depth:
\begin{equation}
\label{deqn_ex19}
\begin{array}{c}{z={\frac{f_{x}\cdot B}{d}}}\end{array}
\end{equation} Where $f_x$ is the normalized focal length, and $B$ is the distance between the optical centers of the two cameras, called the baseline distance.

\section{Experiments}

We conducted multiple experiments to evaluate the performance of our system on four popular publicly available datasets, namely KITTI 
\cite{geiger2013vision}, EuRoC 
\cite{burri2016euroc}, TUM 
\cite{sturm2012benchmark} and 4Season 
\cite{wenzel20214seasons} and actual sequences captured in real campus scenes. These include challenging night scenes with low light and strong light variations to evaluate and validate the robustness and accuracy of \Ours~system in adapting to environments with different light variations at different times of the day.

We performed all experiments on a computer equipped with an Intel Core i5-12400F CPU and a GeForce RTX 3060, and the deep network uses officially provided pre-trained models. Note that there are many essential parameters in the open-source framework of the original ORB-SLAM2, including the image pyramid model-related variables in the feature extraction matching, camera frame rate, depth threshold, etc. To ensure fairness, we test our system on different dataset sequences using configuration parameters consistent with the original ORB-SLAM2. This step also demonstrates the vital portability of our system.

\subsection{Datasets}
Table \ref{tab:table_1_datasets} shows the main characteristics of the dataset used in our experiments. We categorized the dataset into two main classes based on the lighting conditions of different sequential scenes and conducted experimental tests accordingly. We qualitatively and quantitatively evaluated the robustness and accuracy of the system in adapting to the varying lighting conditions over different periods within a day.
\begin{table*}[t]
\caption{MAIN ASPECTS RELATED TO THE PRESENTED BENCHMARK DATASETS
\label{tab:table_1_datasets}}
\centering
\begin{threeparttable} 
\setlength{\tabcolsep}{2mm} 
\renewcommand{\arraystretch}{1.8} 
\begin{tabular}{|l|c|c|c|c|c|l|}
\hline
\multicolumn{1}{|c|}{Dataset} & \multicolumn{1}{c|}{\begin{tabular}[c]{@{}c@{}}Scene \\[-2mm] Type\end{tabular}} & \multicolumn{1}{c|}{\begin{tabular}[c]{@{}c@{}}Scene \\[-2mm]  Characteristics\end{tabular}} & \multicolumn{1}{c|}{\begin{tabular}[c]{@{}c@{}}Time \\[-2mm]  Period\end{tabular}} & \multicolumn{1}{c|}{\begin{tabular}[c]{@{}c@{}}Lighting \\[-2mm]  Condition\end{tabular}} & \multicolumn{1}{c|}{\begin{tabular}[c]{@{}c@{}}Sequence \\[-2mm]  Amount\end{tabular}} & \multicolumn{1}{c|}{Sample} \\  \hline
\multirow{2}{*}{KITTI}  & Indoor  & \multicolumn{1}{c|}{-}  & \multicolumn{1}{c|}{-}   & \multicolumn{1}{c|}{-}  &\multicolumn{1}{c|}{-}  & \multirow{2}{*}{\includegraphics[width=0.35\columnwidth,height=\textheight,keepaspectratio]{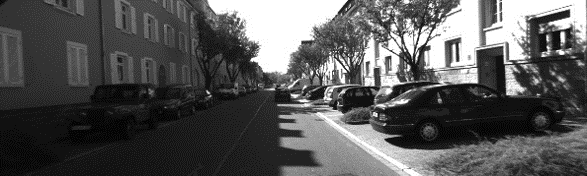}}
        \\ \cline{2-6}   & Outdoor  & Urban streets  & Afternoon & Bright & 11 &  \\ \hline
\multirow{3}{*}{EuRoC}    & \multirow{2}{*}{Indoor\tnote{*}}  & Room   & Morning    & Normal    & 6     & \multirow{3}{*}{\includegraphics[width=0.35\columnwidth,height=\textheight,keepaspectratio]{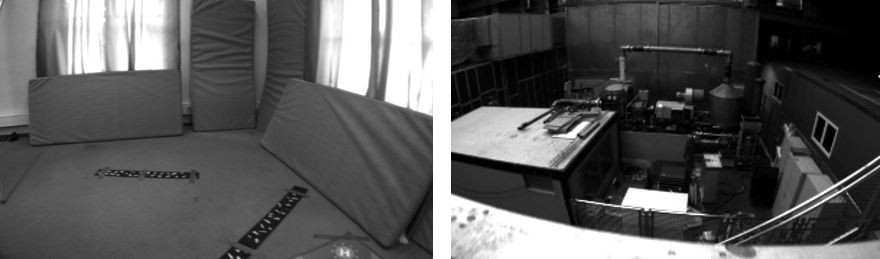}}      \\ \cline{3-6}    &    & Large factory  & Evening    & Dusky    & 5   &   \\ \cline{2-6} 
 & Outdoor    & \multicolumn{1}{c|}{-}    & \multicolumn{1}{c|}{-}    & \multicolumn{1}{c|}{-}   &\multicolumn{1}{c|}{-}    &    \\ \hline
\multirow{2}{*}{TUM RGB-D}    & Indoor  & Office   & Afternoon  & Well-lit     & 7    & \multirow{2}{*}{\includegraphics[width=0.35\columnwidth,height=\textheight,keepaspectratio]{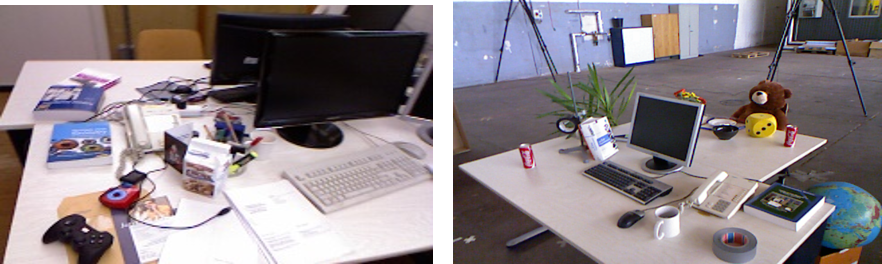}}            \\ \cline{2-6} & Outdoor  & \multicolumn{1}{c|}{-}    & \multicolumn{1}{c|}{-}  & \multicolumn{1}{c|}{-}   &\multicolumn{1}{c|}{-}   &  \\ \hline
\multirow{2}{*}{4Season}   & Indoor  & Parking garage   & Evening   & Dimly-lit  & 1 & \multirow{2}{*}{\includegraphics[width=0.35\columnwidth,height=\textheight,keepaspectratio]{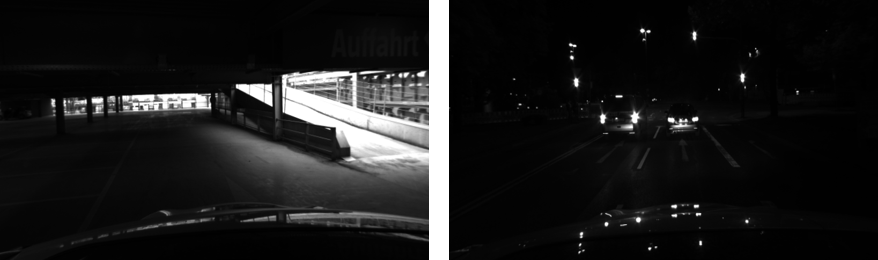}}   \\ \cline{2-6}  & Outdoor  & City streets   & Night   & Dark    & 5   &  \\ \hline
\multirow{2}{*}{Real-World}   & Indoor & Gallery    & Night   & Dimly-lit  & 1  & \multirow{2}{*}{\includegraphics[width=0.35\columnwidth,height=\textheight,keepaspectratio]{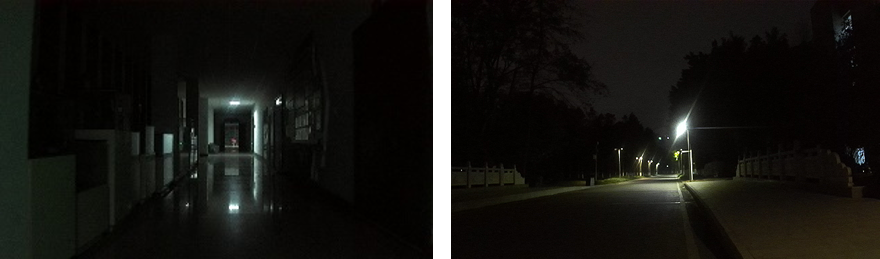}}  \\ \cline{2-6}  & Outdoor  & Campus roads   & Night  & Dark    & 2    &   \\ \hline 
\end{tabular}
\begin{tablenotes} 
		\item[*] The EuRoC dataset contains two types of indoor sequences: normally lit rooms and a dusky large factory. 
     \end{tablenotes}
\end{threeparttable}
\end{table*}

\subsubsection{\bf{Well-lit Scene Datasets}}

We perform extensive experimental tests on three types of datasets, namely KITTI, EuRoC, and TUM, which cover indoor and outdoor scenes with good lighting conditions. To fully validate the overall performance of \Ours~system on the mentioned benchmark datasets, which are most commonly used to evaluate VO/VSLAM algorithms.

The KITTI dataset is a sequence of 22 stereo images of city and highway environments recorded under normal daytime lighting conditions. We only use 11 sequences from 00 to 10 because they are the only stereo sequences that provide ground truth. We will evaluate monocular and binocular vision system performance on this dataset.

The EuRoC dataset contains 11 stereo sequences ranging from slow flight in good visual conditions to fast flight in motion blur and low light conditions. Recordings were taken by a micro air vehicle (MAV) in two rooms and a sizeable industrial scene. Since most of the room sequences had good lighting conditions, the factory scene lacked sufficient lighting. Therefore, we divided this dataset into two categories for separate experimental tests. The room scene (sequences from V1\_01 to V2\_03) is the Well-lit Scene, and the large factory scene (sequences from MH\_01 to MH\_05) is the Dark Scene.

The TUM RGB-D dataset contains several room sequences from RGB-D sensor recordings, most taken with a handheld camera, to evaluate object reconstruction and visual SLAM/odometry methods under different texture, illumination, and structure conditions. From these, we selected sequences for testing under good lighting conditions, typically used by most RGB-D methods for evaluation.

\subsubsection{\bf{Dark Scene Datasets}}

Currently, visual SLAM systems represented by ORB-SLAM2 face great difficulties maintaining robustness in complex environments under low-light conditions. Therefore, to better validate the accurate localization performance of our method in low-light scenes, we choose the public 4Season dataset and the actual sequences captured in natural campus scenes for testing.

\begin{figure}[h]
\centering
\subfloat[]{\includegraphics[width=\linewidth]{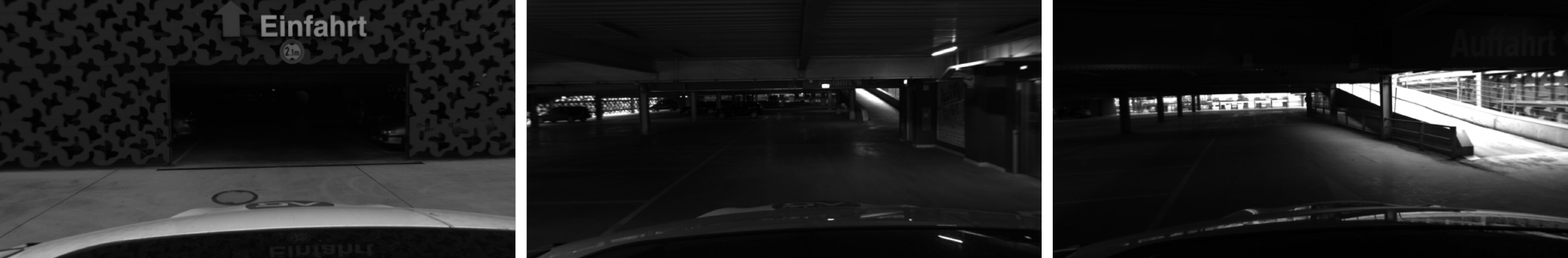}%
\label{4Season_1}}
\hfil
\subfloat[]{\includegraphics[width=\linewidth]{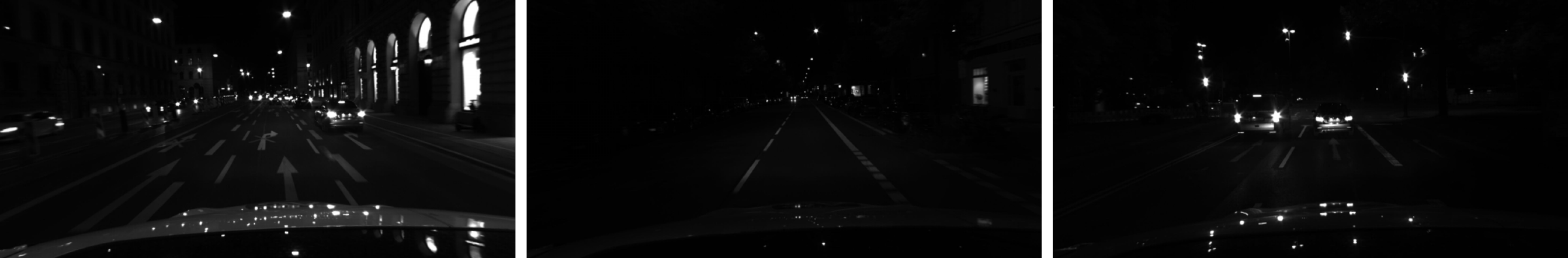}%
\label{4Season_2}}
\caption{Some scenes of parking lots and night roads selected from the 4Season dataset. (a) Parking Garage. (b) Old Town.}
\label{fig_6_4season_sample}
\end{figure}

As a novel and challenging large-scale dataset, the 4Season dataset covers a variety of scenes with different seasonal variations, weather, and lighting conditions that cause drastic changes in appearance. The motion sequences that are low-light scenes and have specific trajectory truth values are parking\_2021-05-10\_19-15-19 and oldtown\_2021-05-10\_21-32-00. The former is a multi-level (three-story) parking garage scene that includes most of the low-light indoor parking garage scenes and some outdoor scenes. The latter is an urban street scene shot at 21:32, where the entire street segment is in a completely dark environment with only some weak local lights, which poses a great challenge to the vision-based SLAM algorithm. The sequence has 24658 image frames with a total length of 5713 m. To test and evaluate the performance of the proposed system under low-light conditions more intuitively, we selected five road segments with different scene characteristics from this long sequence and then compared them with the ORB-SLAM2 system. Fig \ref{fig_6_4season_sample} shows some scenes of parking lots and night streets in the selected paths.

\begin{figure}[h]
\centering
\includegraphics[width=\linewidth]{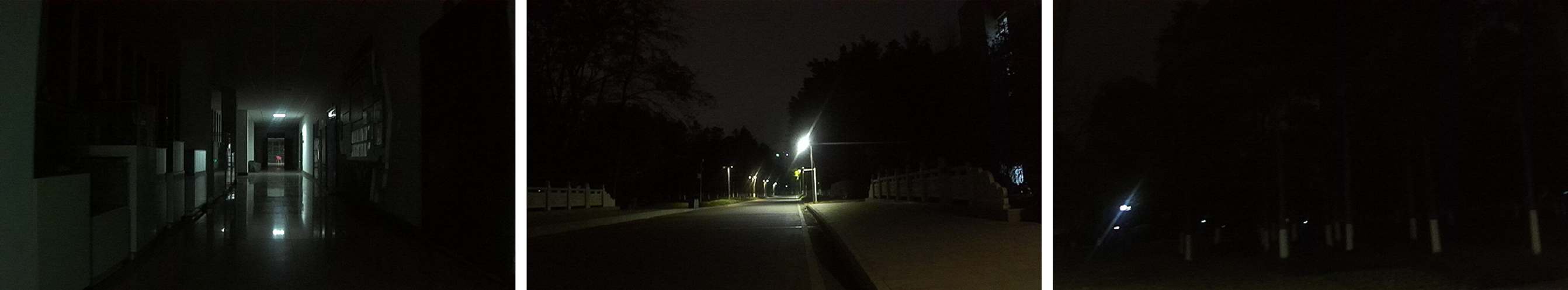}
\caption{Some real scenes are collected from the campus environment. The picture on the left shows an indoor corridor, the middle shows an outdoor campus road, and the right shows a forest scene beside the road.}
\label{fig_7_mydataset_sample}
\end{figure}

To further investigate the localization effect of the \Ours~system in actual scenes, we conducted the acquisition of real scene sequences in the campus environment. The data recording equipment was a ZED 2i stereo camera and a Livox HAP solid-state LiDAR with ground truth provided by the VoxelMap++ 
\cite{wu2023voxelmap++} method. We collect a total of three different paths. Figure \ref{fig_7_mydataset_sample} presents some sample scenarios for these sequences. Sequence 01 shows a corridor scene in an indoor laboratory building at 01:55 AM, with only some weak lighting. Sequences 02 and 03 are outdoor roads shot at 22:23, with a few streetlights on both sides and a primarily dark forest surrounding the environment. The above sequences pose a great challenge to the localization task of the visual SLAM system, which meets our requirement of evaluating the system for testing in very challenging scenes with low light or even dark conditions.

\subsection{Trajectory Evaluation}\label{sec:Trajectory_Evaluation}

We perform quantitative and qualitative comparisons between each dataset sequence's estimated trajectories and the ground truth data. To accurately assess the positioning accuracy of the systems, we adopted the EVO method 
\cite{grupp2017evo}, which compares the trajectory results of the different systems on each path in detail using the absolute translational root mean square error (RMSE) as a measure, thus visualizing the differences in their respective positioning accuracies. Among them, the best results are shown in bold black, while "X" indicates that the method fails to track the entire path in a complete run. A "-" means that no data is available.

\subsubsection{{Well-lit Scene Datasets}}

\paragraph{KITTI Dataset}

\begin{figure}[h]
\centering
\includegraphics[width=\linewidth]{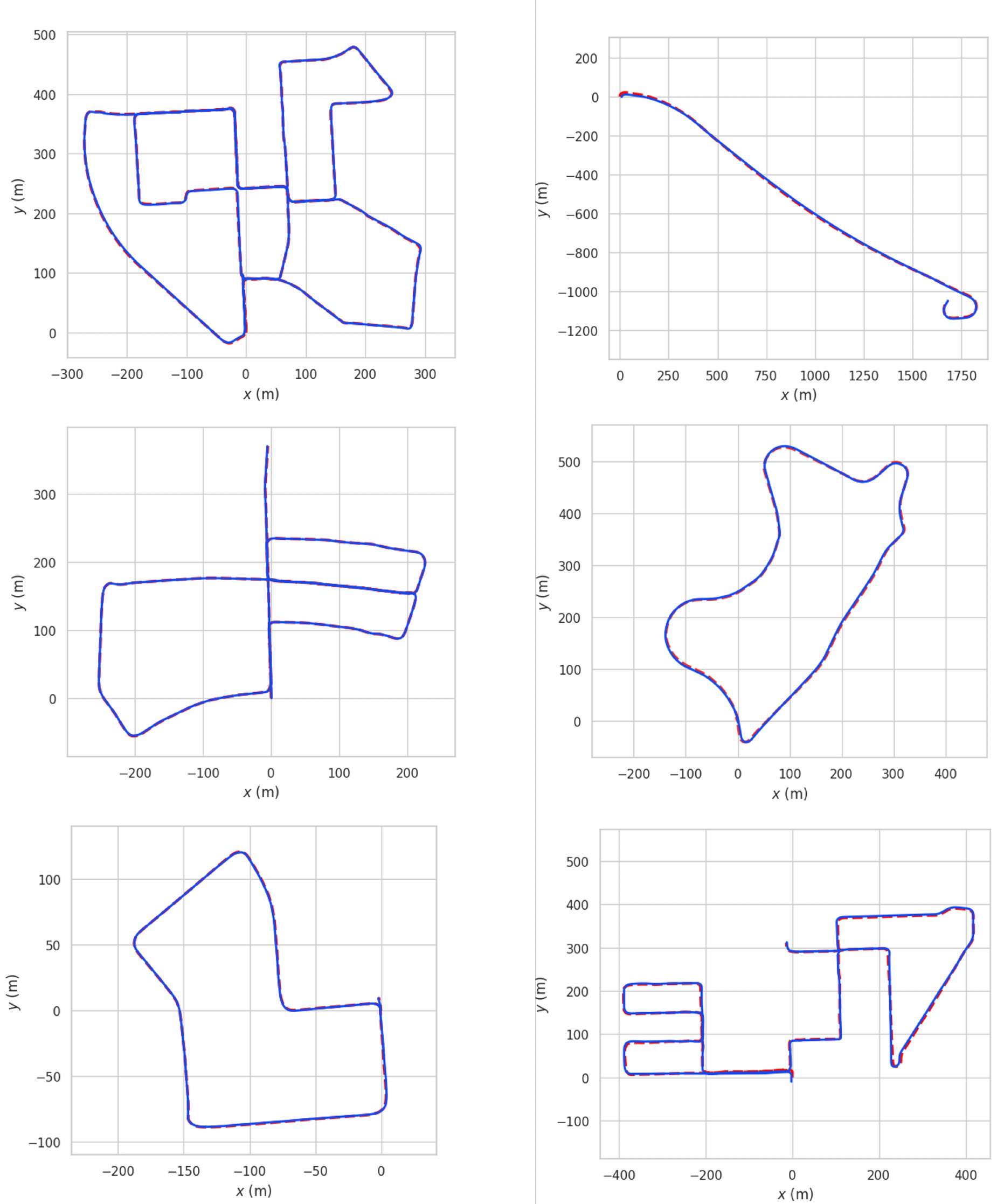}
\caption{Estimated trajectory (blue) and ground truth (red) in KITTI 00, 01, 05, 07, 08, 09.}
\label{fig_8_kitti_stereo}
\end{figure}

Table \ref{tab:table_2_kitti_stereo} shows the quantitative comparisons of binocular vision data results on several KITTI datasets. We compare with the current state-of-the-art methods ORB-SLAM2 and Stereo LSD-SLAM 
\cite{engel2015large}, and the experimental results show that our method achieves better results in most sequences. Fig \ref{fig_8_kitti_stereo} shows an example of estimated trajectories compared to ground truth trajectories for some sequences.

\begin{table}[h]
\caption{COMPARISION OF TRANSLATION RMSE (m) IN THE KITTI DATASET
\label{tab:table_2_kitti_stereo}}
\centering
\setlength{\tabcolsep}{2.3mm} 
\renewcommand{\arraystretch}{1.2} 
\begin{tabular}{lccc}
\specialrule{0.5pt}{0pt}{2pt} 
\hline
Sequence & Ours          & \begin{tabular}[c]{@{}c@{}}ORB-SLAM2\\[-1mm]    (Stereo)\end{tabular} & \begin{tabular}[c]{@{}c@{}}Stereo \\[-1mm]    LSD-SLAM\end{tabular} \\ \hline
00       & \textbf{0.97} & 1.30                                                            & 1.00                                                          \\
01       & \textbf{6.85} & 10.40                                                           & 9.00                                                          \\
02       & 5.19          & 5.70                                                            & \textbf{2.60}                                                 \\
03       & \textbf{0.35} & 0.60                                                            & 1.20                                                          \\
04       & \textbf{0.13} & 0.20                                                            & 0.20                                                          \\
05       & \textbf{0.38} & 0.80                                                            & 1.50                                                          \\
06       & \textbf{0.46} & 0.80                                                            & 1.30                                                          \\
07       & \textbf{0.38} & 0.50                                                            & 0.50                                                          \\
08       & \textbf{3.18} & 3.60                                                            & 3.90                                                          \\
09       & \textbf{2.28} & 3.20                                                            & 5.60                                                          \\
10       & 1.10 & \textbf{1.00}                                                   & 1.50                                                          \\ 
\specialrule{0.5pt}{0pt}{2pt} 
\hline
\end{tabular}
\end{table}

Compared to ORB-SLAM2, \Ours~shows significant superiority in localization accuracy for all sequences except sequence 10. Among them, the improvement is more than 0.3 m in most cases (sequences 00, 01, 02, 05, 06, 08, 09), mainly due to many scenes with drastic changes in illumination and low texture in these sequences. Examples include walking through a forest, entering a residential area from an open road, etc. In the challenging highway sequence 01, the system improved the localization accuracy by about 34.1$\%$. Compared with Stereo LSD-SLAM, our system achieves improved localization accuracy on all sequences except sequence 02.

Table \ref{tab:table_3_kitti_mono} shows the quantitative comparisons of monocular vision data results on sequences from the KITTI dataset. We compare with the state-of-the-art popular traditional vision method ORB-SLAM2 and the deep learning-based method SuperPoint-SLAM [11], as well as the method proposed by Sun \etal 
\cite{sun2022improving}. For monocular vision, the KITTI dataset appears very challenging because many sequences contain scenes with cars moving at high speeds (recorded at only 10 fps), rotating rapidly, and with many low-texture regions. Sequence 01 is the only highway sequence in the dataset. Due to the high-speed motion of the vehicles and the resulting low frame rate in the image sequence, monocular visual systems can successfully track very few nearby feature points. Therefore, it is almost impossible to track the camera's motion trajectory fully in this scenario using traditional methods.

\begin{table}[h]
\caption{COMPARISION OF TRANSLATION RMSE (m) IN THE KITTI DATASET
\label{tab:table_3_kitti_mono}}
\centering
\setlength{\tabcolsep}{1.5mm} 
\renewcommand{\arraystretch}{1.2} 
\begin{tabular}{lcccc}
\specialrule{0.5pt}{0pt}{2pt} 
\cline{1-5}
Sequence & Ours            & \begin{tabular}[c]{@{}c@{}}ORB-SLAM2\\[-1mm]    (Mono)\end{tabular} & \begin{tabular}[c]{@{}c@{}}SuperPoint- \\[-1mm]    SLAM\end{tabular} & Sun et al. \cite{sun2022improving} \\ \cline{1-5}
00       & \textbf{6.03}   & 7.45                                                            & X                                                             & 61.79               \\
01       & \textbf{107.08} & X                                                               & X                                                             & X                   \\
02       & \textbf{20.59}           & 31.70                                                           & X                                                    & 28.79               \\
03       & 1.04   & 1.42                                                            & 1.04                                                          & \textbf{0.98}                \\
04       & \textbf{0.85}   & 1.19                                                            & 0.35                                                          & 0.89                \\
05       & 5.34   & 7.02                                                            & \textbf{3.73}                                                          & 31.08               \\
06       & \textbf{13.17}  & 15.62                                                           & 14.27                                                         & 45.67               \\
07       & \textbf{1.32}   & 3.37                                                            & 3.02                                                          & 15.99               \\
08       & \textbf{37.85}  & 51.43                                                           & 39.63                                                         & 42.09               \\
09       & \textbf{45.91}  & 51.85                                                           & X                                                             & 45.95               \\
10       & 7.08   & 7.90                                                   & \textbf{5.31}                                                          & 8.11                \\ 
\specialrule{0.5pt}{0pt}{2pt} 
\cline{1-5}
\end{tabular}
\end{table}

The results show that \Ours~is able to handle all sequences, including the challenging sequence 01 (a car driving at a high-speed scene), while other existing state-of-the-art vision methods fail to track sequence 01. It demonstrates our system's ability to ensure stable feature extraction and data association in highly degraded scenes. Compared to ORB-SLAM2, \Ours~has higher localization accuracy, which is improved by at least 1.42 m or more in most sequences (sequences 00, 01, 02, 05, 06, 07, 08, 09). Among them, sequence 07 contains a loop that can be detected only in the last few frames of the sequence, which significantly impacts the system's performance for global pose graph optimization during loop detection. The proposed system substantially improves localization accuracy compared to ORB-SLAM2, which can improve about 60.8$\%$.

In addition, it is easy to see that the end-to-end deep learning method proposed in 
\cite{sun2022improving} does not perform as well as our proposed method in terms of localization accuracy, and it is less robust in the face of new environments and complexities with large localization errors. Compared with the SuperPoint-SLAM deep learning method based on SuperPoint feature points, \Ours~shows better performance in most sequences. It is easy to see that SuperPoint-SLAM fails to track sequences 00, 01, 02, and 09. The method performs poorly in tracking in low-texture environments with few buildings and dense forests. In contrast, our proposed system can handle these challenging scenes with stable and efficient feature tracking, showing strong robustness and superiority.

\paragraph{EuRoC Dataset}

\begin{figure}[h]
\centering
\includegraphics[width=\linewidth]{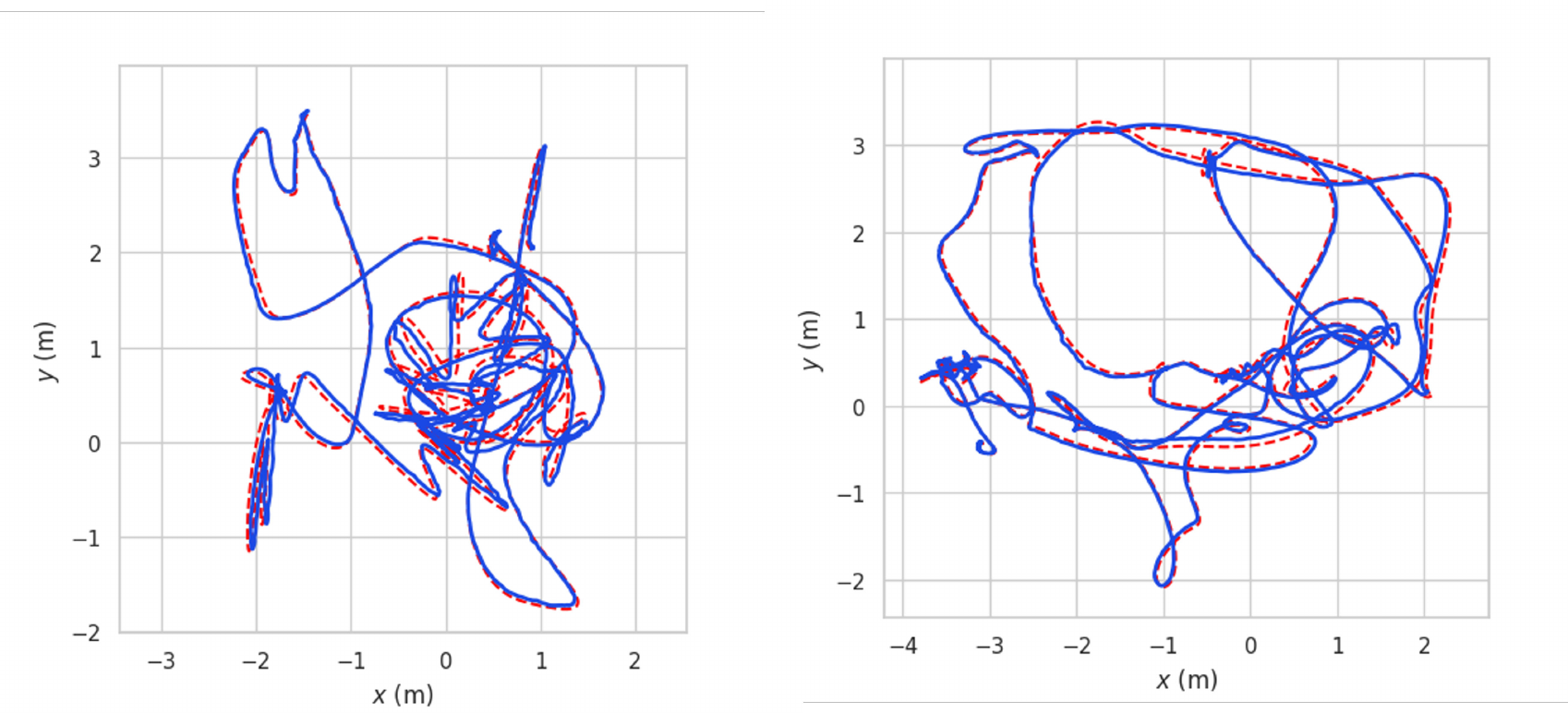}
\caption{Estimated trajectory (blue) and ground truth (red) in EuRoC V1\_03\_difficult and V2\_02\_medium.}
\label{fig_9_euroc_v}
\end{figure}

Table \ref{tab:table_4_euroc_v} shows the quantitative comparison between the algorithms on the indoor sequences of the EuRoC dataset (sequences from V1\_01 to V2\_03). Fig \ref{fig_9_euroc_v} shows an example of the estimated trajectories for some sequences compared to the ground truth trajectories. The experimental data show that although ORB-SLAM2 has achieved sub-centimeter localization accuracy and is generally more accurate than Stereo LSD-SLAM, our proposed method still outperforms ORB-SLAM2 in all sequences. 

\begin{table}[h]
\caption{COMPARISION OF TRANSLATION RMSE (cm) IN THE EUROC DATASET
\label{tab:table_4_euroc_v}}
\centering
\setlength{\tabcolsep}{1.5mm} 
\renewcommand{\arraystretch}{1.2} 
\begin{tabular}{lccc}
\specialrule{0.5pt}{0pt}{2pt} 
\hline
Sequence          & Ours          & \begin{tabular}[c]{@{}c@{}}ORB-SLAM2\\[-1mm] (Stereo)\end{tabular} & \begin{tabular}[c]{@{}c@{}}Stereo\\[-1mm] LSD-SLAM\end{tabular} \\ \hline
V1\_01\_easy      & 8.4           & 8.6                                                             & \textbf{6.6}                                               \\
V1\_02\_medium    & \textbf{5.8}  & 6.4                                                             & 7.4                                                        \\
V1\_03\_difficult & \textbf{6.2}  & 11.6                                                            & 8.9                                                        \\
V2\_01\_easy      & \textbf{6.5}  & 6.7                                                             & -                                                          \\
V2\_02\_medium    & \textbf{5.2}  & 6.3                                                             & -                                                          \\
V2\_03\_difficult & \textbf{12.5} & X                                                               & -                                                          \\ 
\specialrule{0.5pt}{0pt}{2pt} \hline
\end{tabular}
\end{table}

Due to the way MAVs fly and the environment resulting in sequences containing many fast-blurring transformed images and low-textured scenes, the system will be heavily dependent on the performance of the features. We can easily find that \Ours~performs very well. The system improves localization accuracy by at least 0.2 cm in all sequences. And on the challenging, difficult level sequence V1\_03, the proposed system shows strong robustness and accuracy compared to ORB-SLAM2, with a localization accuracy improvement of about 46.6$\%$.

Because of the single texture of the wall with severe motion blur, ORB-SLAM2 fails to track when running the V2\_03\_difficult sequence. However, \Ours~is able to effectively handle this challenging, complex sequence and keep the localization accuracy error around 12.5 cm. It is excellent proof that our method is more robust and accurate.

\paragraph{TUM RGB-D Dataset}

\begin{figure}[h]
\centering
\includegraphics[width=\linewidth]{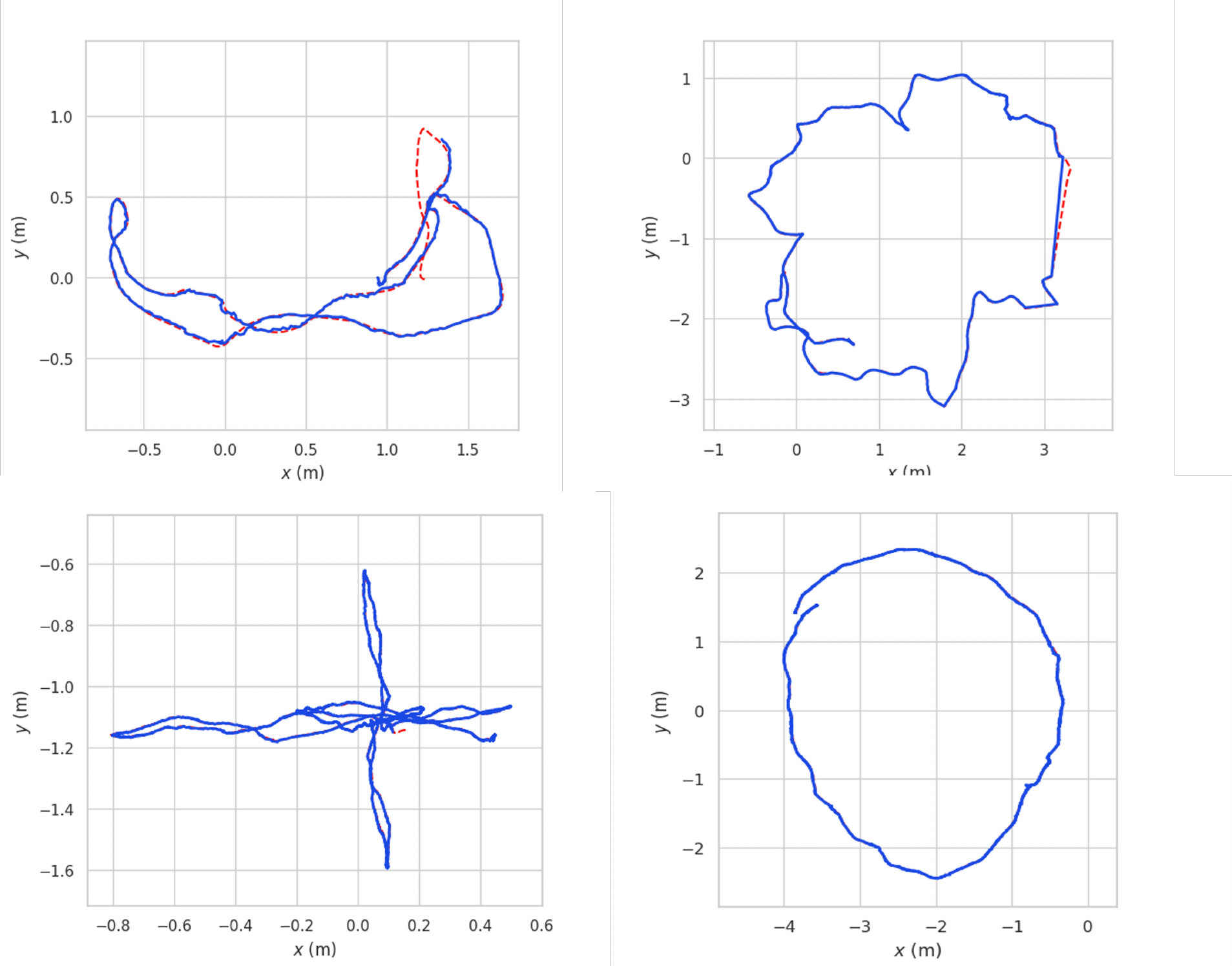}
\caption{Estimated trajectory (blue) and ground truth (red) in TUM RGB-D fr1\_desk, fr2\_desk, fr2\_xyz, fr3\_nst.}
\label{fig_10_tum}
\end{figure}

As shown in Table \ref{tab:table_5_tum}, we compare our method with the following state-of-the-art methods on the TUM RGB-D dataset: ORB-SLAM2, RGBDTAM 
\cite{concha2017rgbdtam}, and ElasticFusion 
\cite{whelan2016elasticfusion}. Fig \ref{fig_10_tum} shows an example of the estimated trajectories for some sequences compared to the ground truth trajectories. The results show that \Ours~outperforms the other methods on all sequences except the fr3/office sequence, where the accuracy is similar to ORB-SLAM2. Among them, on the fr3/nst sequence, the localization accuracy is improved by about 31.6$\%$ compared to the ORB-SLAM2 method.

\begin{table}[h]
\caption{COMPARISION OF TRANSLATION RMSE (cm) IN THE TUM RGB-D DATASET
\label{tab:table_5_tum}}
\centering
\setlength{\tabcolsep}{1.5mm} 
\renewcommand{\arraystretch}{1.2} 
\begin{tabular}{lcccc}
\specialrule{0.5pt}{0pt}{2pt} 
\hline
Sequence   & Ours         & \begin{tabular}[c]{@{}c@{}}ORB-SLAM2\\[-1mm]    (RGB-D)\end{tabular} & RGBDTAM & \begin{tabular}[c]{@{}c@{}}Elastic\\[-1mm]    Fusion\end{tabular} \\ \hline
fr1/desk   & \textbf{1.4} & 1.6                                                            & 2.7     & 2.0                                                            \\
fr1/desk2  & \textbf{2.1} & 2.2                                                            & 4.2     & 4.8                                                            \\
fr1/room   & \textbf{4.2} & 4.7                                                            & 15.5    & 6.8                                                            \\
fr2/desk   & \textbf{0.8} & 0.9                                                            & 2.7     & 7.1                                                            \\
fr2/xyz    & \textbf{0.3} & 0.4                                                            & 0.4     & 1.1                                                            \\
fr3/office & 1.0          & 1.0                                                            & 2.7     & 1.7                                                            \\
fr3/nst    & \textbf{1.3} & 1.9                                                            & 1.6     & 1.6                                                            \\ 
\specialrule{0.5pt}{0pt}{2pt} 
\hline
\end{tabular}
\end{table}

\subsubsection{{Dark Scene Datasets}}

\paragraph{EuRoC Dataset}

\begin{figure}[h]
\centering
\includegraphics[width=\linewidth]{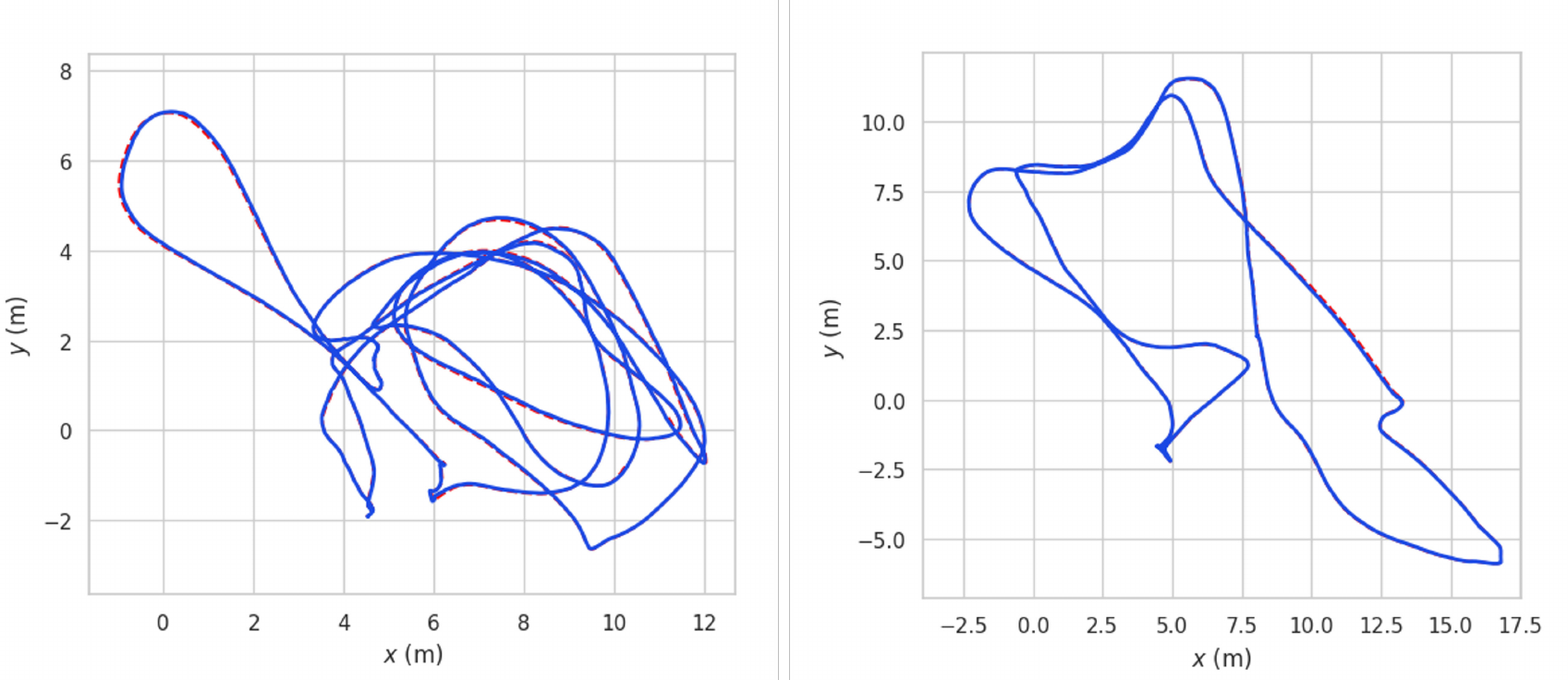}
\caption{Estimated trajectory (blue) and ground truth (red) in EuRoC  MH\_03\_medium and MH\_05\_difficult.}
\label{fig_11_euroc_mh}
\end{figure}

Table \ref{tab:table_6_euroc_mh} shows a quantitative comparison between the algorithms on the EuRoC dataset large plant sequences (sequences from MH\_01 to MH\_05). Fig \ref{fig_11_euroc_mh} shows an example of the estimated trajectories compared to the ground truth trajectories for some sequences. Since Stereo LSD-SLAM does not give available experimental data in the above sequences, we compare it with the most popular ORB-SLAM2 method.

\begin{table}[h]
\caption{COMPARISION OF TRANSLATION RMSE (cm) IN THE TUM RGB-D DATASET
\label{tab:table_6_euroc_mh}}
\centering
\setlength{\tabcolsep}{3.9mm} 
\renewcommand{\arraystretch}{1.2} 
\begin{tabular}{lcc}
\specialrule{0.5pt}{0pt}{2pt}
\hline
Sequence          & Ours          & \begin{tabular}[c]{@{}c@{}}ORB-SLAM2\\[-1mm]    (RGB-D)\end{tabular} \\ \hline
MH\_01\_easy      & 3.5           & 3.5                                                            \\
MH\_02\_easy      & \textbf{3.7}  & 4.0                                                            \\
MH\_03\_medium   & \textbf{3.3}  & 3.9                                                            \\
MH\_04\_difficult & \textbf{11.2} & 11.5                                                           \\
MH\_05\_difficult & \textbf{3.9}  & 6.0                                                            \\ \specialrule{0.5pt}{0pt}{2pt} \hline
\end{tabular}
\end{table}

The data show that our proposed method outperforms ORB-SLAM2 in all sequences except the MH\_01\_easy (with the same accuracy as ORB-SLAM2). Regarding localization accuracy, \Ours~improves by at least 0.3 cm or more in most sequences. Moreover, on the challenging, difficult grade sequence MH\_05, the proposed system shows strong robustness and accuracy compared to ORB-SLAM2, with an improvement of about 35.0$\%$ in localization accuracy. There are many consecutive low-light scenes in this sequence, and the system will rely heavily on the performance of image features. However, traditional manual feature-based methods have difficulty tracking a sufficient number and accuracy of features, which is a severe challenge to the localization task of visual SLAM systems. These data show the significant performance improvement that \Ours~brings when dealing with low-light scenes.

\begin{figure*}[!t]
\centering
\subfloat[]{\includegraphics[width=0.33\linewidth]{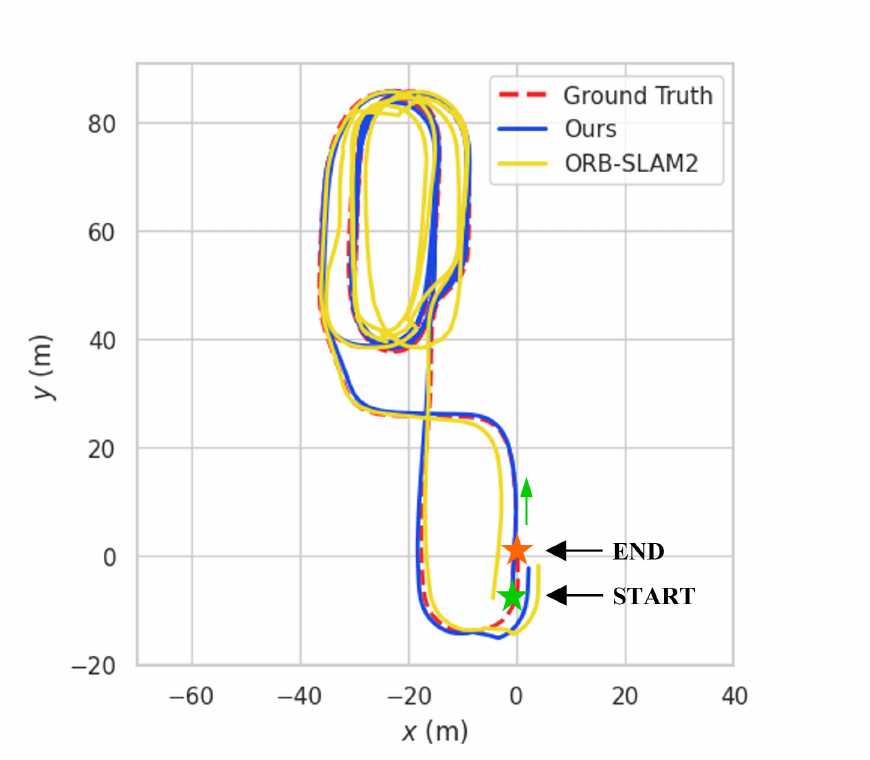}%
\label{4Season_01}}
\subfloat[]{\includegraphics[width=0.33\linewidth]{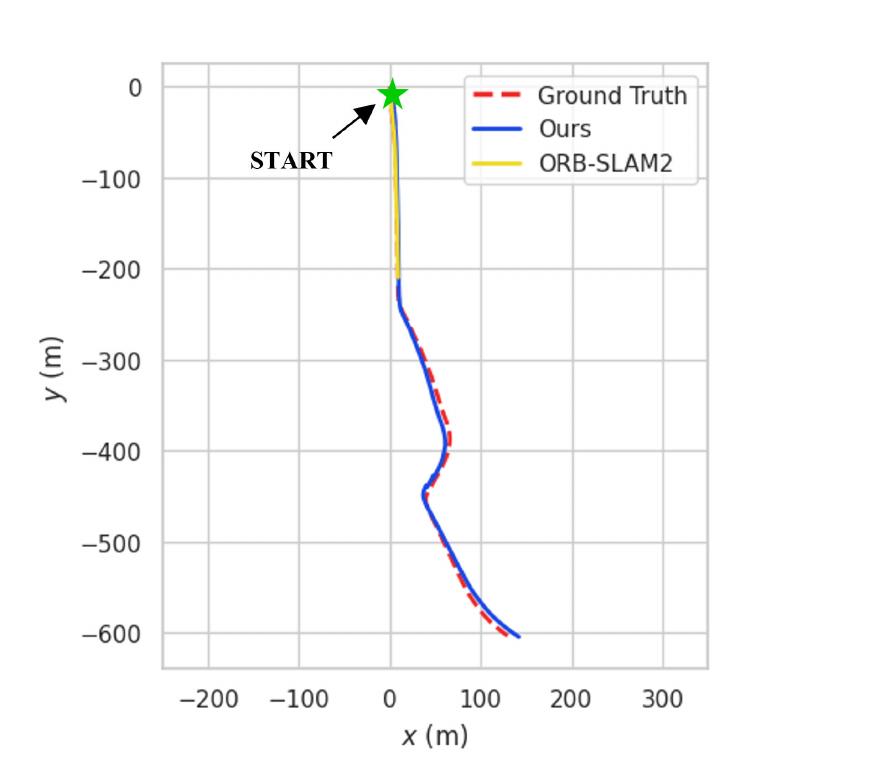}%
\label{4Season_02}}
\subfloat[]{\includegraphics[width=0.33\linewidth]{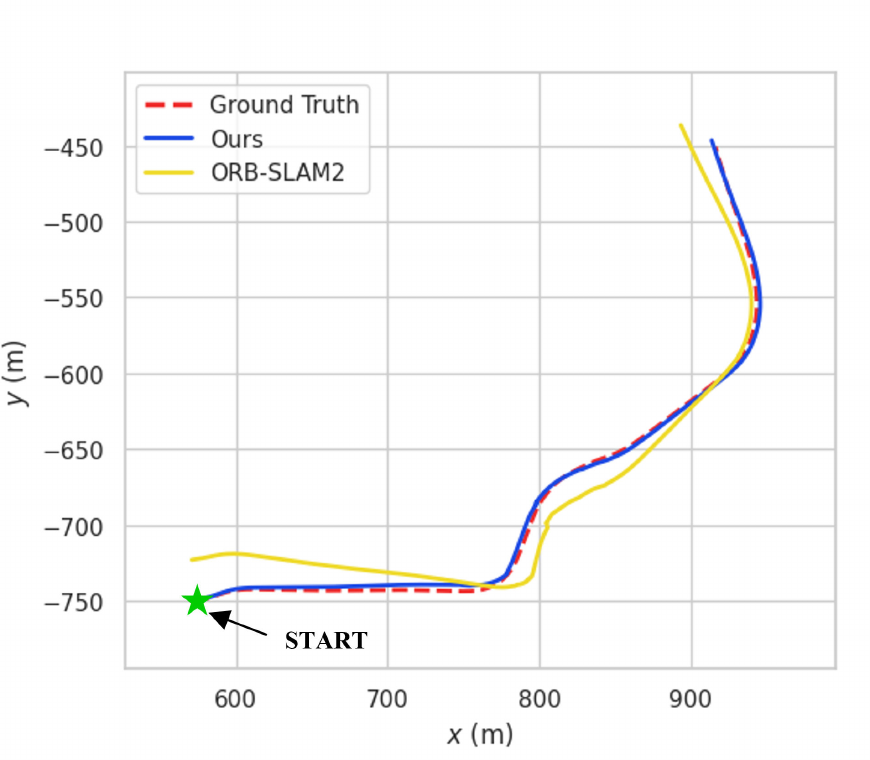}%
\label{4Season_03}}
\\
\subfloat[]{\includegraphics[width=0.33\linewidth]{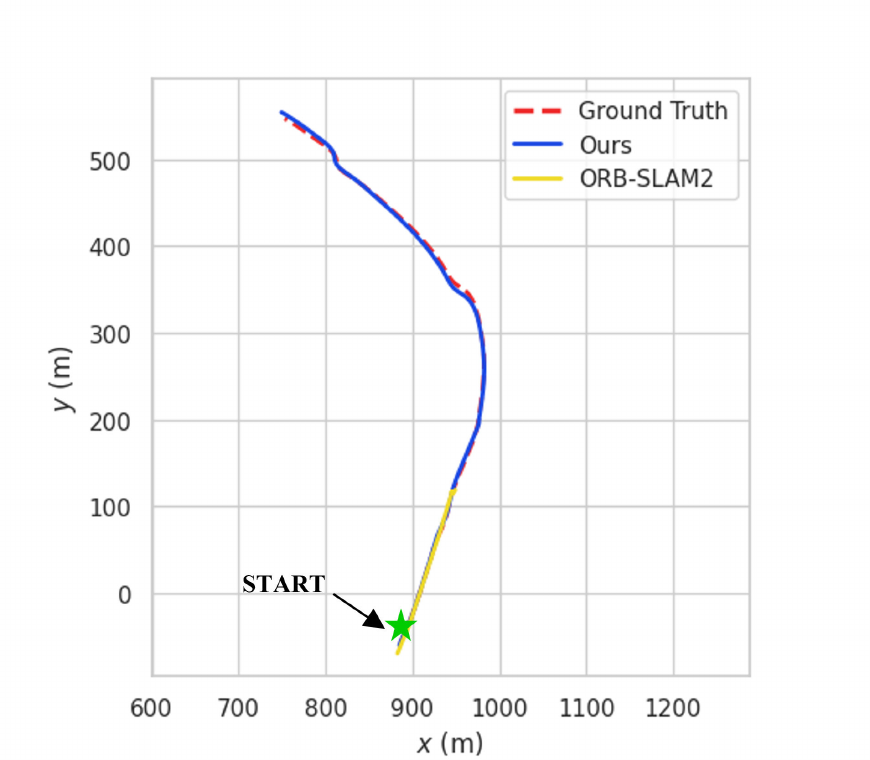}%
\label{4Season_04}}
\subfloat[]{\includegraphics[width=0.33\linewidth]{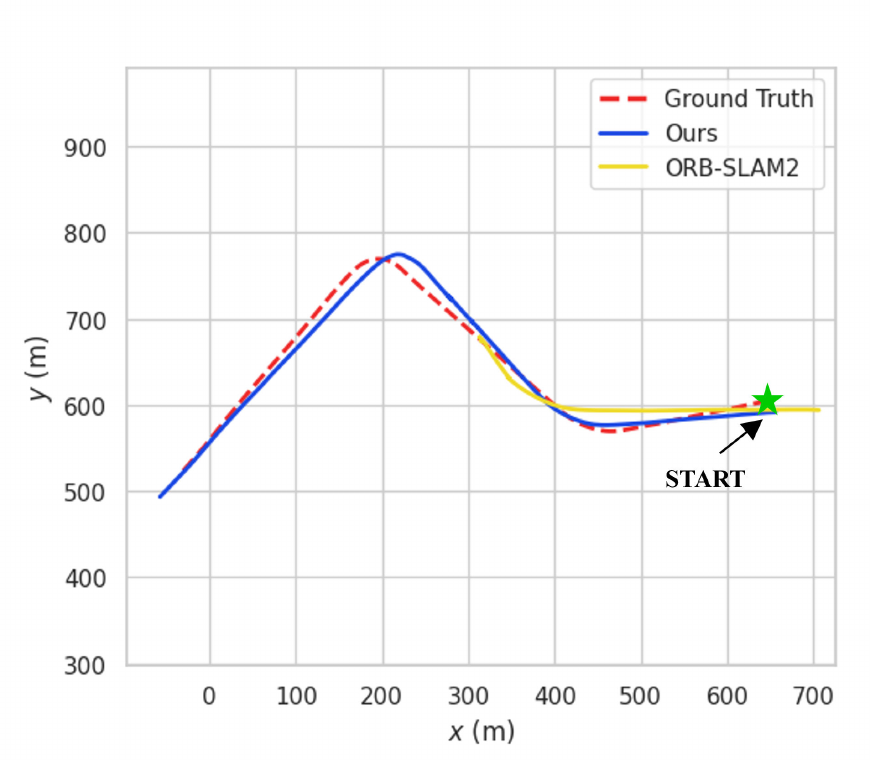}%
\label{4Season_05}}
\subfloat[]{\includegraphics[width=0.33\linewidth]{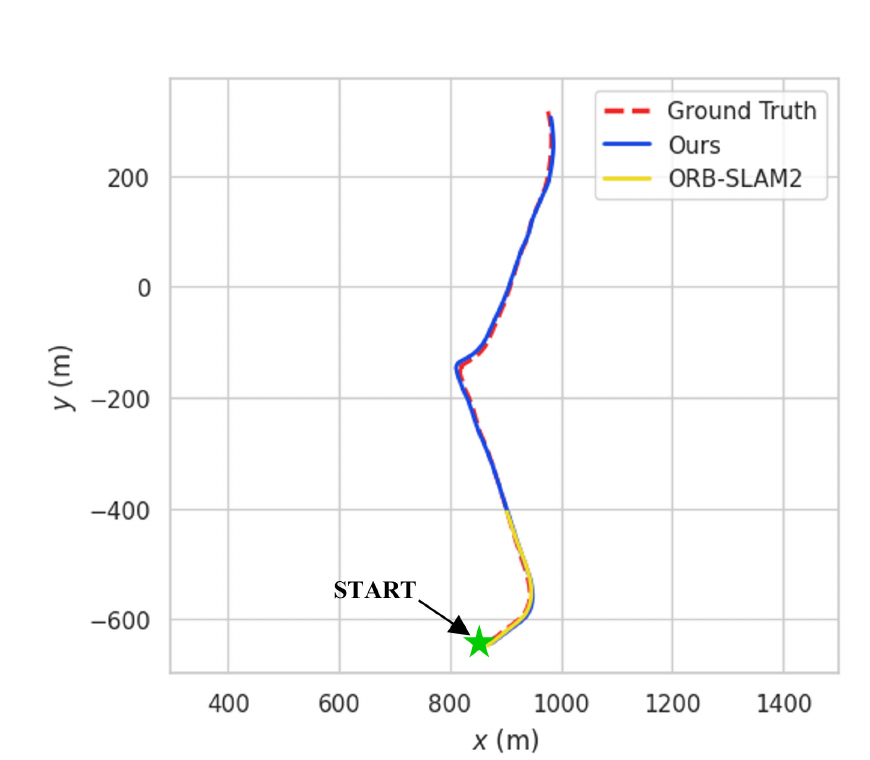}%
\label{4Season_06}}
\caption{Comparison of trajectories between ORB-SLAM2 (yellow) and our method (blue) on six 4Season Dataset sequences. (a) Parking Garage. (b) OT\_01\_easy. (c) OT\_02\_medium. (d) OT\_03\_medium. (e) OT\_04\_difficult. (f) OT\_05\_difficult. Mark the starting point with a green five-pointed star. Also, mark the endpoint in (a) with an orange five-pointed star and indicate the initial direction of motion with a green arrow.}
\label{fig_12_4season}
\end{figure*}

\paragraph{4Season Dataset}

Table \ref{tab:table_7_4season} shows the results of our approach compared to ORB-SLAM2 on the 4Season dataset sequence. Fig \ref{fig_12_4season} illustrates the results of the corresponding estimated trajectories of the two methods compared to the ground truth trajectory accuracy. The experimental results show that \Ours~is less affected by the nighttime low-light environment on most test paths, exhibits strong robustness, and achieves higher localization accuracy than the popular visual ORB-SLAM2 method. However, ORB-SLAM2 faces great difficulties adapting to the scene under these low-light conditions. In low-light environments, the details and contrasts in the images are drastically reduced, and the traditional manual features become very difficult to extract and match, leading to direct tracking failures when the system is in operation.

\begin{table}[h]
\caption{COMPARISION OF TRANSLATION RMSE (m) IN THE 4SEASON DATASET
\label{tab:table_7_4season}}
\centering
\setlength{\tabcolsep}{1.2mm} 
\renewcommand{\arraystretch}{1.2} 
\begin{tabular}{llcc}
\specialrule{0.5pt}{0pt}{2pt}
\hline
Sequence           & \multicolumn{1}{c}{Scene Characteristics}                                                 & Ours            & \begin{tabular}[c]{@{}c@{}}ORB-SLAM2\\[-1mm]    (Stereo)\end{tabular} \\ \hline
Parking Garage     & \begin{tabular}[c]{@{}l@{}}Dramatic light \\[-1mm] changes\end{tabular}                                                                   & \textbf{0.826}  & 2.502                                                           \\
OT\_01\_easy      & \begin{tabular}[c]{@{}l@{}}Reduced lighting \\[-1mm] at the corner\end{tabular}                 & \textbf{10.161} & X                                                               \\
OT\_02\_medium     & \begin{tabular}[c]{@{}l@{}}Low lighting and \\[-1mm] continuous turns\end{tabular}              & \textbf{4.659}  & 17.351                                                          \\
OT\_03\_medium    & \begin{tabular}[c]{@{}l@{}}Sudden darkening \\[-1mm] of lighting\end{tabular}                   & \textbf{7.503}  & X                                                               \\
OT\_04\_difficult  & \begin{tabular}[c]{@{}l@{}}Limited streetlights and\\[-1mm] few structured objects\end{tabular} & \textbf{25.202} & X                                                               \\
OT\_05\_difficult & \begin{tabular}[c]{@{}l@{}}Significantly degraded\\[-1mm] lighting environment\end{tabular}     & \textbf{7.734}  & X                                                               \\ \specialrule{0.5pt}{0pt}{2pt} \hline
\end{tabular}
\end{table}

In the Parking Garage sequence, most of the indoor scenes are dimly lit, and there is the effect of drastic light changes when switching between indoor and outdoor scenes, and the visual SLAM system can quickly lose track. Based on the experimental results, it is easy to see that despite the help of the relocalization function, the ORB-SLAM2 system still has a large error in localization accuracy, and the estimated trajectory has a significant offset on the whole. In contrast, \Ours~achieves a higher accuracy with an improvement of about 67.0$\%$ compared to ORB-SLAM2, which reflects the superiority of the proposed system under low-light conditions. 

In the nighttime Old Town street sequences, ORB-SLAM2 almost wholly fails. When certain scenes in the path lack illumination, such as in the OT\_01\_easy sequence, the system immediately fails to track. In the OT\_02\_medium sequence, ORB-SLAM2 was able to complete the sequence due to the many street lights in the path. However, due to the presence of large and continuous turning sections, a sizeable cumulative error is generated. Our method, on the other hand, shows exciting accuracy, improving the localization accuracy by about 73.1$\%$ compared to ORB-SLAM2. In the sequence OT\_03\_medium, the ORB-SLAM2 system loses tracking due to the drastic change in illumination as the car passes through the intersection, and the light suddenly dims. In addition, when there are no structurally rich buildings in the low-light night scene, the system also has difficulty tracking the exact feature points, and it is easy to lose track, such as in sequences OT\_04\_difficult and OT\_05\_difficult, where the system is unable to complete the entire path. Overall, \Ours~provides a more robust and accurate localization solution when adapting to low light environments at night.

\begin{figure*}[!t]
\centering
\subfloat[]{\includegraphics[width=0.33\linewidth]{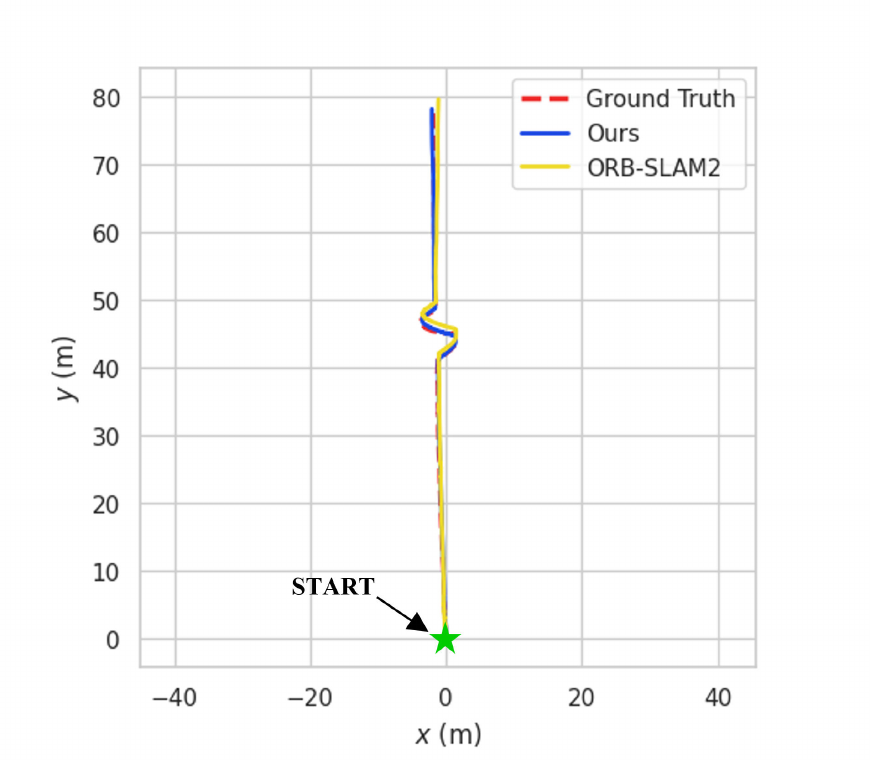}%
\label{5MyDataset_01}}
\subfloat[]{\includegraphics[width=0.33\linewidth]{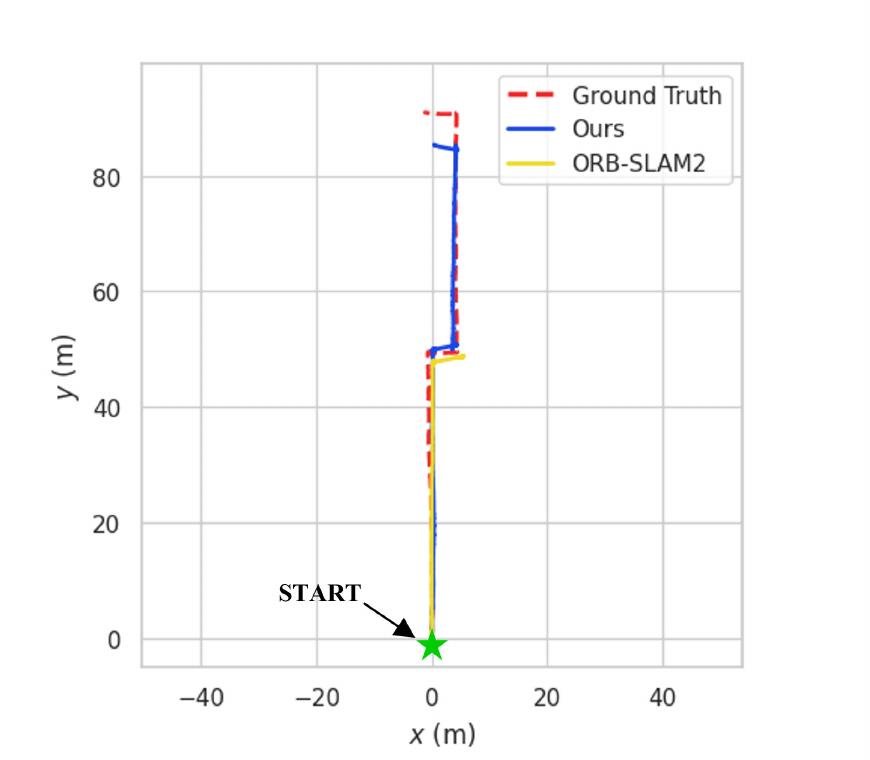}%
\label{5MyDataset_02}}
\subfloat[]{\includegraphics[width=0.33\linewidth]{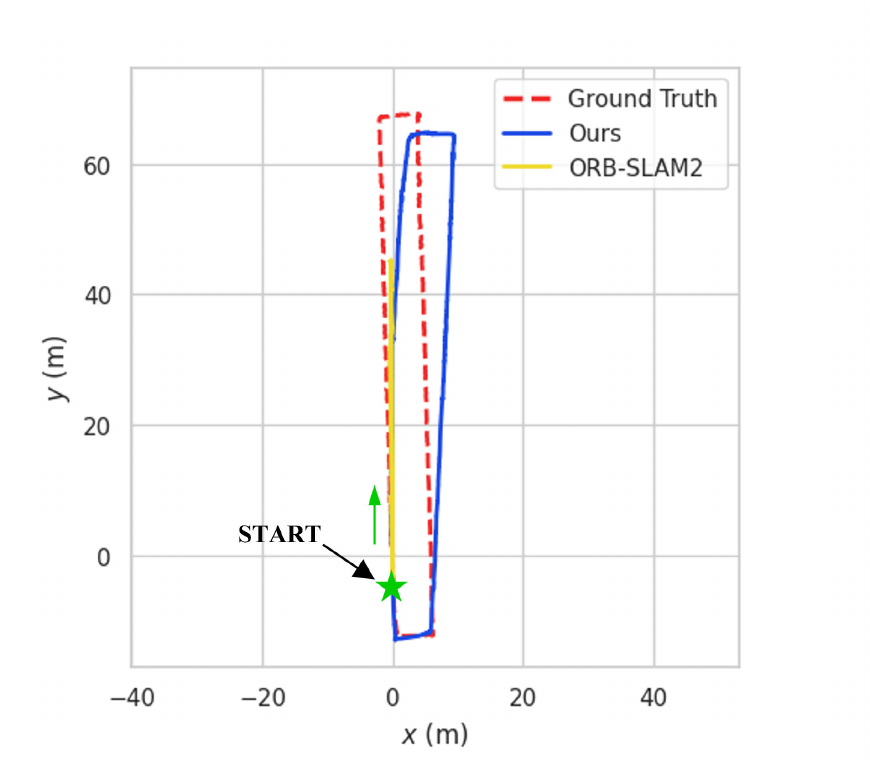}%
\label{5MyDataset_03}}
\caption{Comparison of trajectories between ORB-SLAM2 (yellow) and our method (blue) on My Dataset sequences. (a) Path01. (b) Path02. (c) Path03. Mark the starting point with a green five-pointed star and indicate the initial direction of motion with a green arrow.}
\label{fig_13_mydataset}
\end{figure*}

\paragraph{Real-World Dataset}

Fig \ref{fig_13_mydataset} shows the results of empirical tests comparing the estimated trajectories between the algorithms on actual scene sequences with ground truth trajectories. The results show that ORB-SLAM2 succeeds in tracking only in the indoor Path01 and fails in the other two outdoor road sequences, Path02 and Path03. 

In Path01, the estimated trajectory of ORB-SLAM2 inevitably produces an overall bias due to a low-light scene with continuous large turns. And our method is closer to the ground truth data. In Path02, when our motion path was kept straight from the left side of the road and then turned right to the right side of the road, the camera captured a scene with more woods than streetlights, and the lighting was very dim, causing ORB-SLAM2 to fail in direct tracking. Our method still ensures stable feature tracking over the entire path in this scene with severely degraded lighting conditions. However, the system suffers from scaling errors because there are fewer near-feature points to capture on the right side of the path and more densely clustered distant features. In Path03, the lighting conditions of this scene are even worse, and the left side of the road is an entirely black environment. The system cannot acquire enough effective information, and ORB-SLAM2 cannot adapt to this environment. However, \Ours~is able to complete the entire path, albeit with some deviation in the overall estimated trajectory. It also shows that the proposed system remains robust in night-light scenes.

\subsection{Processing Time}

We used an Intel Core i5-12400F CPU and a GeForce RTX 3060 computer for the evaluation. To evaluate the running speed of our proposed system, we counted the running time of each image frame from the system's input to the output of the pose of that image frame as the elapsed time to process that image frame.

Table \ref{tab:table_8_time} shows the results of testing the statistics of the average running time per frame of different algorithms on KITTI sequence 04. We compare the popular traditional visual SLAM methods (e.g., ORB-SLAM2), hybrid methods based on deep feature descriptors (e.g., SuperPoint-SLAM and GCNv2), and end-to-end methods based on deep learning (e.g., the method proposed by Sun \etal). ORB-SLAM2 relies on the traditional hand-designed ORB features, which are relatively fast in computation but less robust when facing complex environments. SuperPoint-SLAM system is time-consuming regarding network inference, which takes about 0.5 seconds to process a frame and is far from real-time requirements. We cannot obtain the exact time of GCNv2 because it cannot complete the tracking in a sequence. The method proposed by Sun \etal~takes up a lot of computational resources and time in the inference phase of the depth estimation network, which becomes a bottleneck that prevents the system from realizing high processing speed.

\begin{table}[h]
\caption{AVERAGE PROCESSING TIME FOR EACH IMAGE ON KITTI SEQUENCE 04
\label{tab:table_8_time}}
\centering
\setlength{\tabcolsep}{1.2mm} 
\renewcommand{\arraystretch}{1.4} 
\begin{tabular}{cccccc}
\specialrule{0.5pt}{0pt}{2pt}
\hline
                    & Ours & ORB-SLAM2   & \begin{tabular}[c]{@{}c@{}}SuperPoint-\\[-4.5mm]    \\ SLAM\end{tabular} & GCNv2 & Sun et al. \\ \hline
\begin{tabular}[c]{@{}c@{}}Time per\\[-4mm]    \\ Frame (ms)\end{tabular} & 95   & 25 & 513                  & -     & 268        \\ \specialrule{0.5pt}{0pt}{2pt} \hline
\end{tabular}
\end{table}

With the improvement of the algorithm of our proposed parallel image pyramid model, we have reduced the average running time of each image frame from about 337 ms to 95 ms, which effectively improves the overall processing speed and meets the requirement of the system to be able to run in real-time on GPU. In addition, as the robustness and accuracy of the system continue to improve, there is still a lot of room for us to improve the running speed of the whole system.

\section{Conclusion}
We propose a novel \Ours~system that successfully applies deep learning networks to traditional visual SLAM techniques to achieve robust and highly accurate real-time localization at different times of day and under challenging lighting conditions. \Ours~takes full advantage of deep learning. It uses deep localized feature descriptors to replace the traditional hand-designed features. It achieves precise matching between features quickly through a more accurate and efficient deep network. We have demonstrated excellent performance in many experiments.

Experimental results show that \Ours~outperforms traditional hand-crafted feature-based and deep learning-based methods in terms of accuracy and robustness over most dataset sequences. Even in challenging low-light scenes, it can overcome the difficulties that popular algorithms such as ORB-SLAM2 may not operate stably. Thus, the proposed method provides a more robust and accurate localization solution when adapting to different environments.

In future work, we aim to improve the robustness of the proposed system further in more extreme degradation scenes to solve the problem of dealing with complex and challenging localization and map building.


\bibliographystyle{IEEEtran}
\bibliography{myslam}

\begin{IEEEbiography}
[{\includegraphics[width=1in,height=1.25in,clip,keepaspectratio]{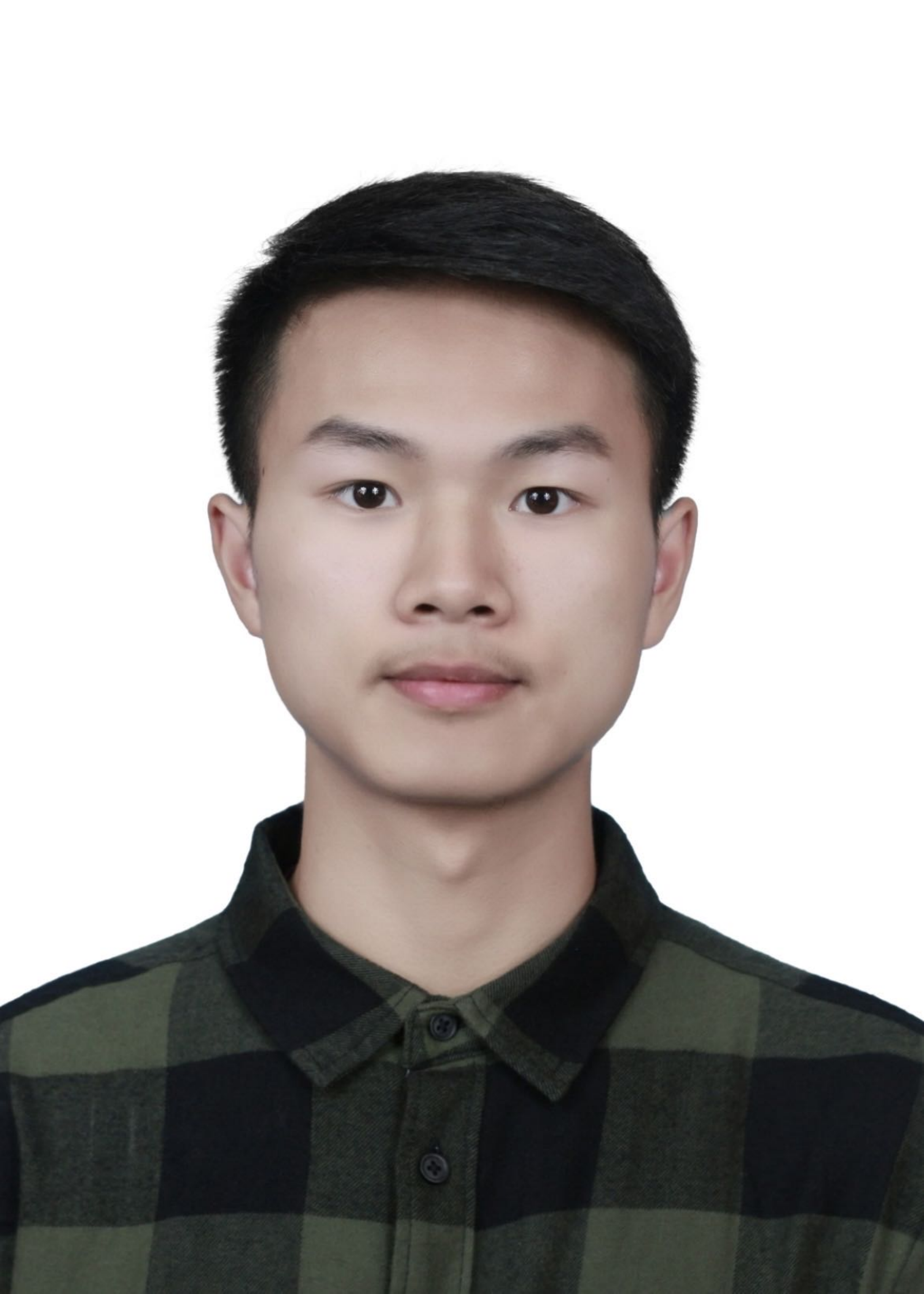}}]{Zhiqi Zhao}
received a Bachelor’s degree in communication engineering from the University of Electronic Science and Technology of China in 2022. He is pursuing a master’s in electronic information at the University of Electronic Science and Technology. His main research interests are visual SLAM, multi-sensor fusion localization, and deep learning methods.
\end{IEEEbiography}

\begin{IEEEbiography}
[{\includegraphics[width=1in,height=1.25in,clip,keepaspectratio]{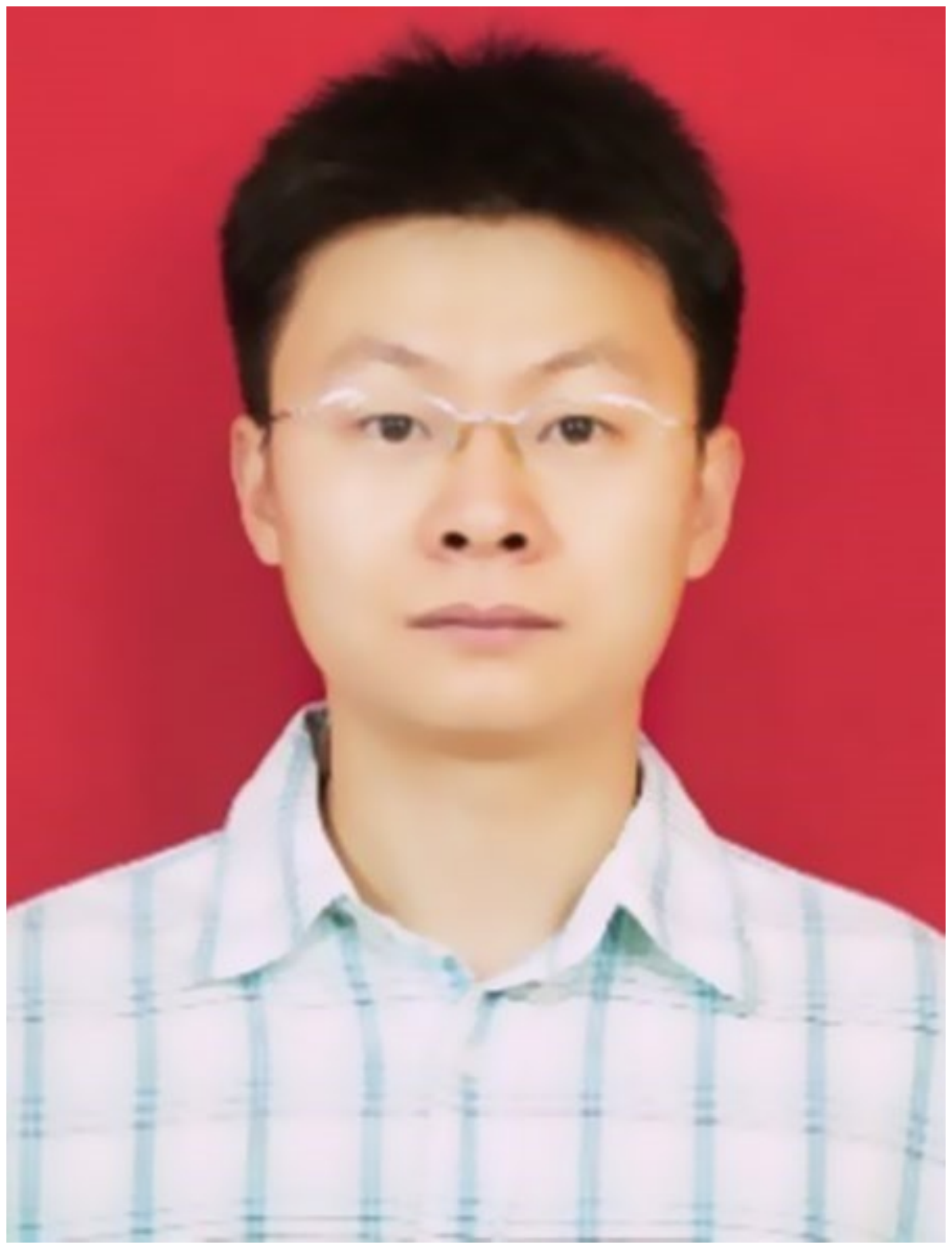}}]{Chang Wu}
received his PhD in University of Electronic Science and Technology of China (UESTC), 2008. He is associate professor in School of information and communication, UESTC. His research fields include indoor location-based service, navigation, communication theory, information technology, channel coding and 5G mobile communication system.
\end{IEEEbiography}

\begin{IEEEbiography}
[{\includegraphics[width=1in,height=1.25in,clip,keepaspectratio]{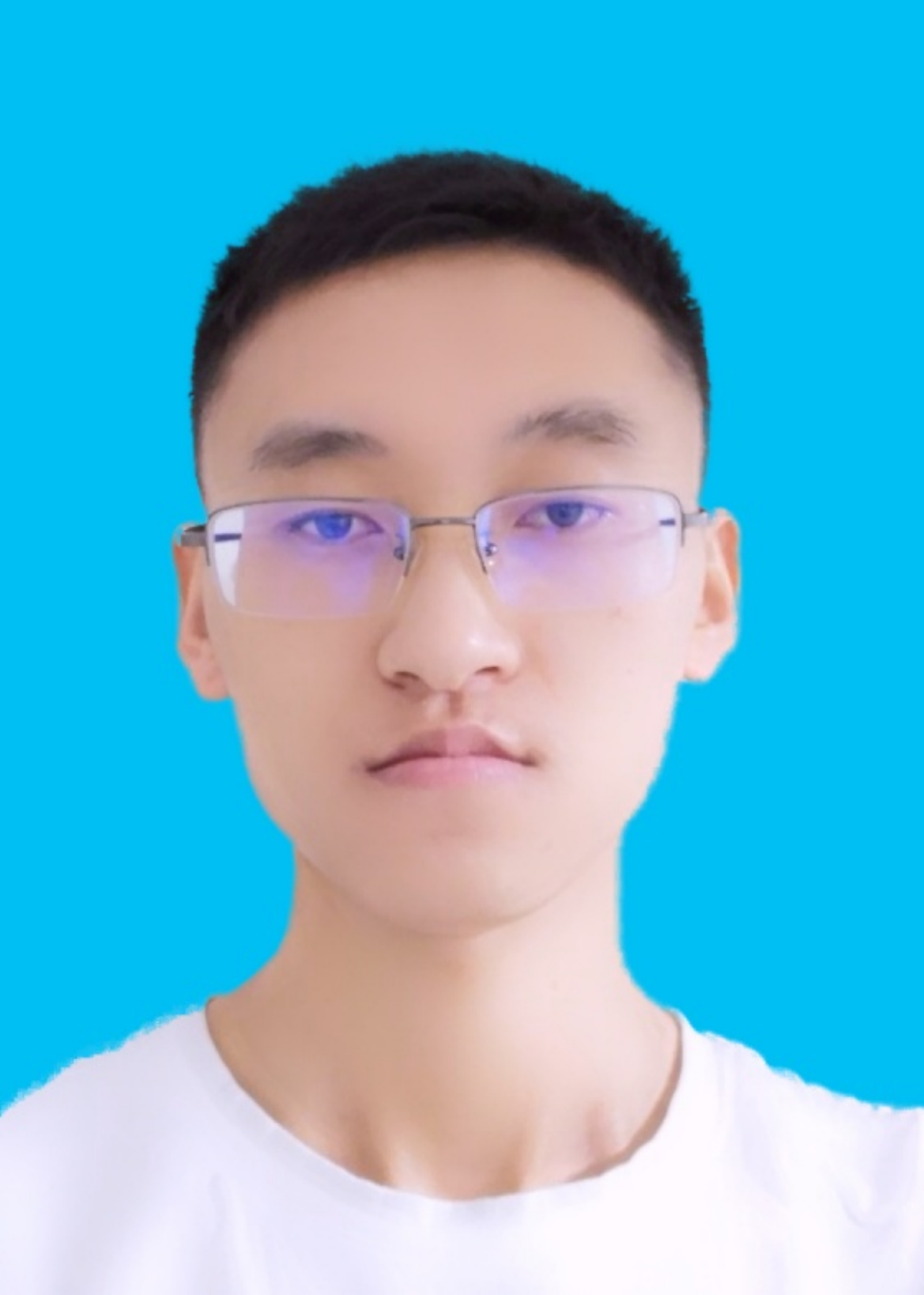}}]{Xiaotong Kong}
received the Bachelor’s degree in communication engineering from University of Electronic Science and Technology of China, in 2021. He is currently pursuing a master’s degree in electronic information at the University of Electronic Science and Technology, where his main research interests are in indoor positioning methods.
\end{IEEEbiography}

\begin{IEEEbiography}
[{\includegraphics[width=1in,height=1.25in,clip,keepaspectratio]{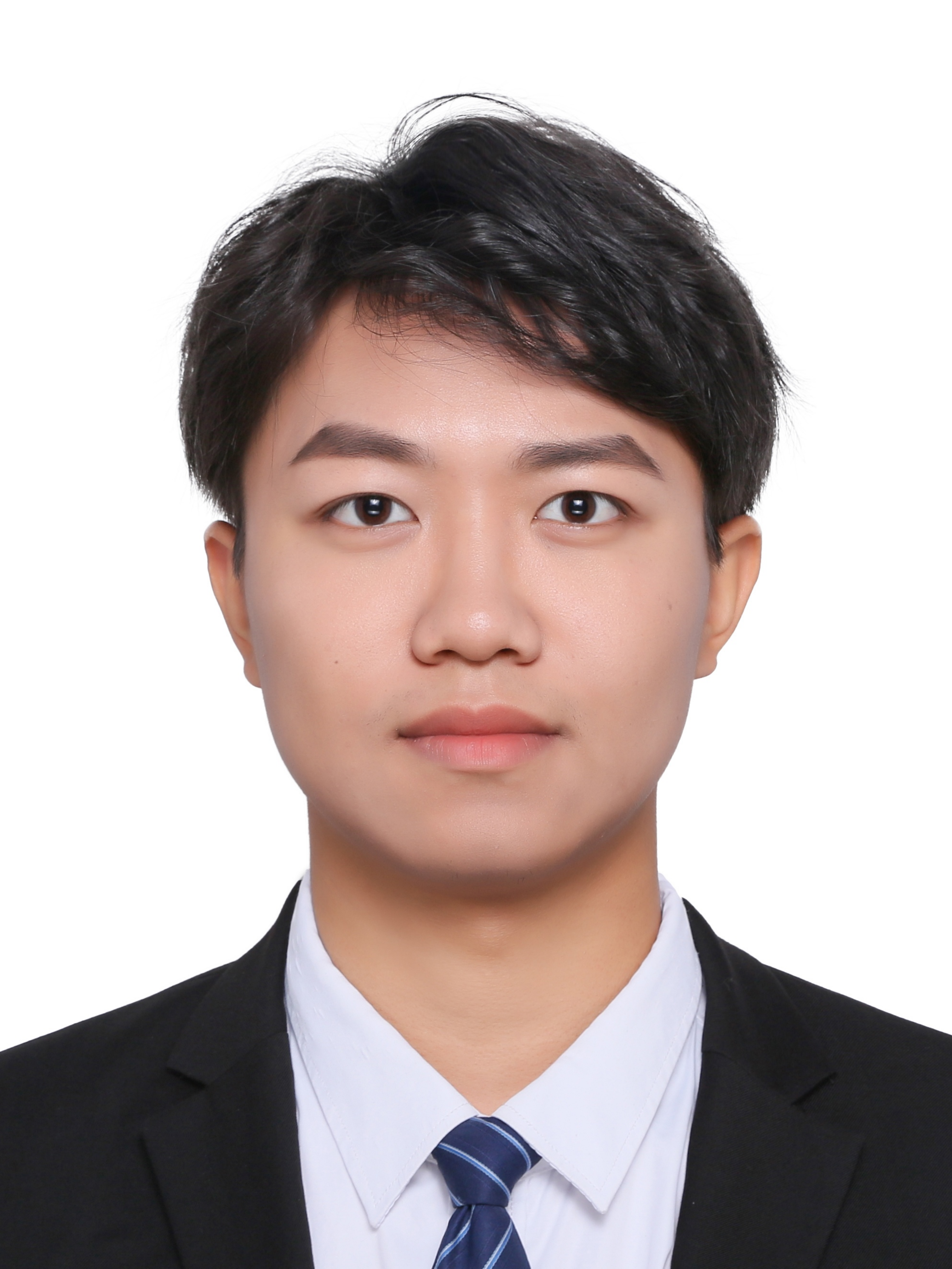}}]{Zejie Lv}
received his bachelor’s degree in electronic information engineering from the University of Electronic Science and Technology of China, Chengdu, China, in 2022 and is currently pursuing a master’s degree in electronic information engineering. His current research includes lidar SLAM algorithm and multi-sensor fusion positioning.
\end{IEEEbiography}

\begin{IEEEbiography}
[{\includegraphics[width=1in,height=1.25in,clip,keepaspectratio]{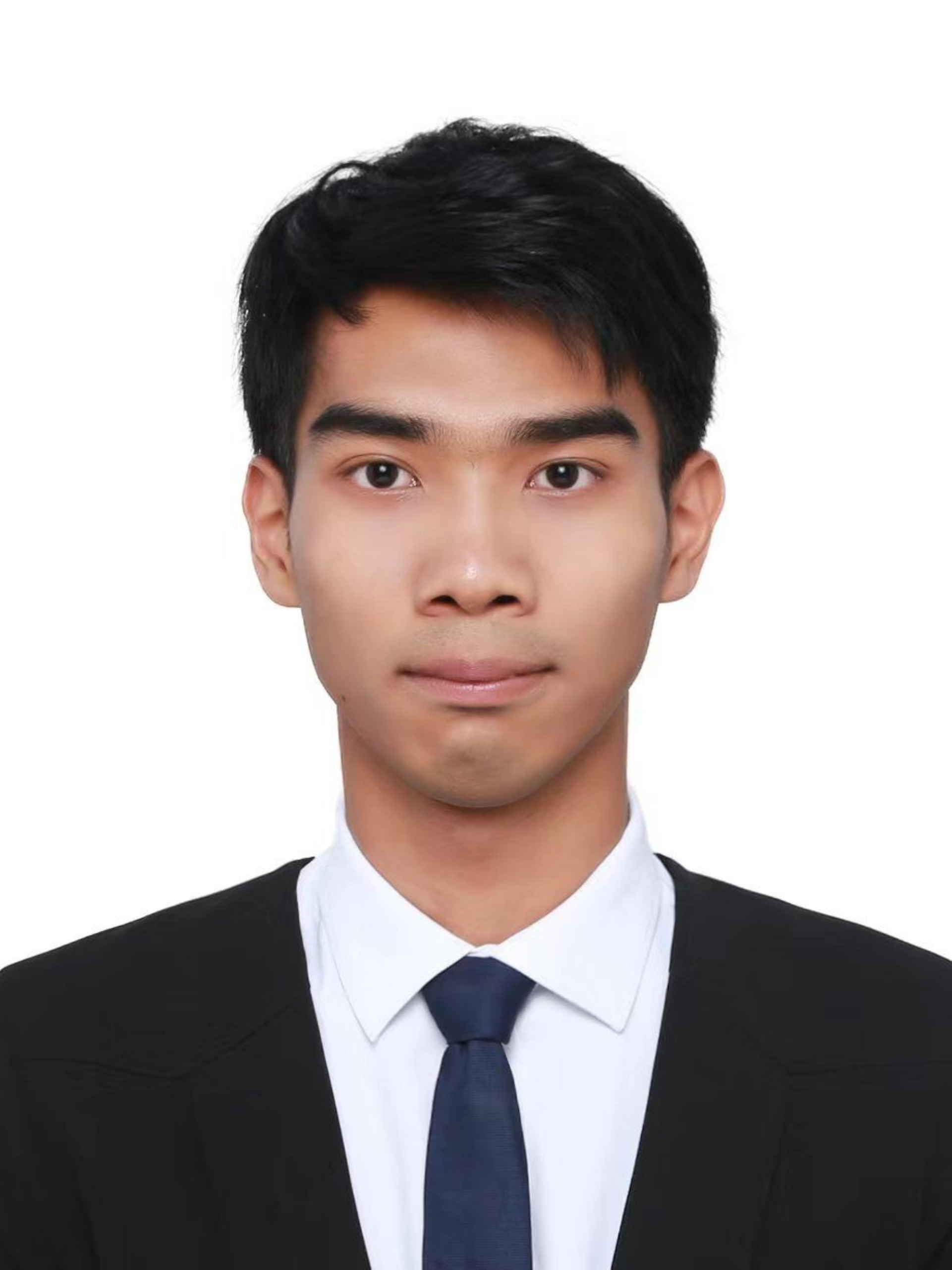}}]{Xiaoqi Du}
received the bachelor’s degree in communication engineering from the University of Electronic Science and Technology of China, Chengdu, China, in 2023, where he is currently pursuing the master’s degree in electronic information engineering. He has experience working in industrial program development and extensive skills in Python, Pytorch and Paddlepaddle. His research interests include computer vision, indoor localization, navigation and quantitative investment combined with the application of machine learning.
\end{IEEEbiography}

\begin{IEEEbiography}
[{\includegraphics[width=1in,height=1.25in,clip,keepaspectratio]{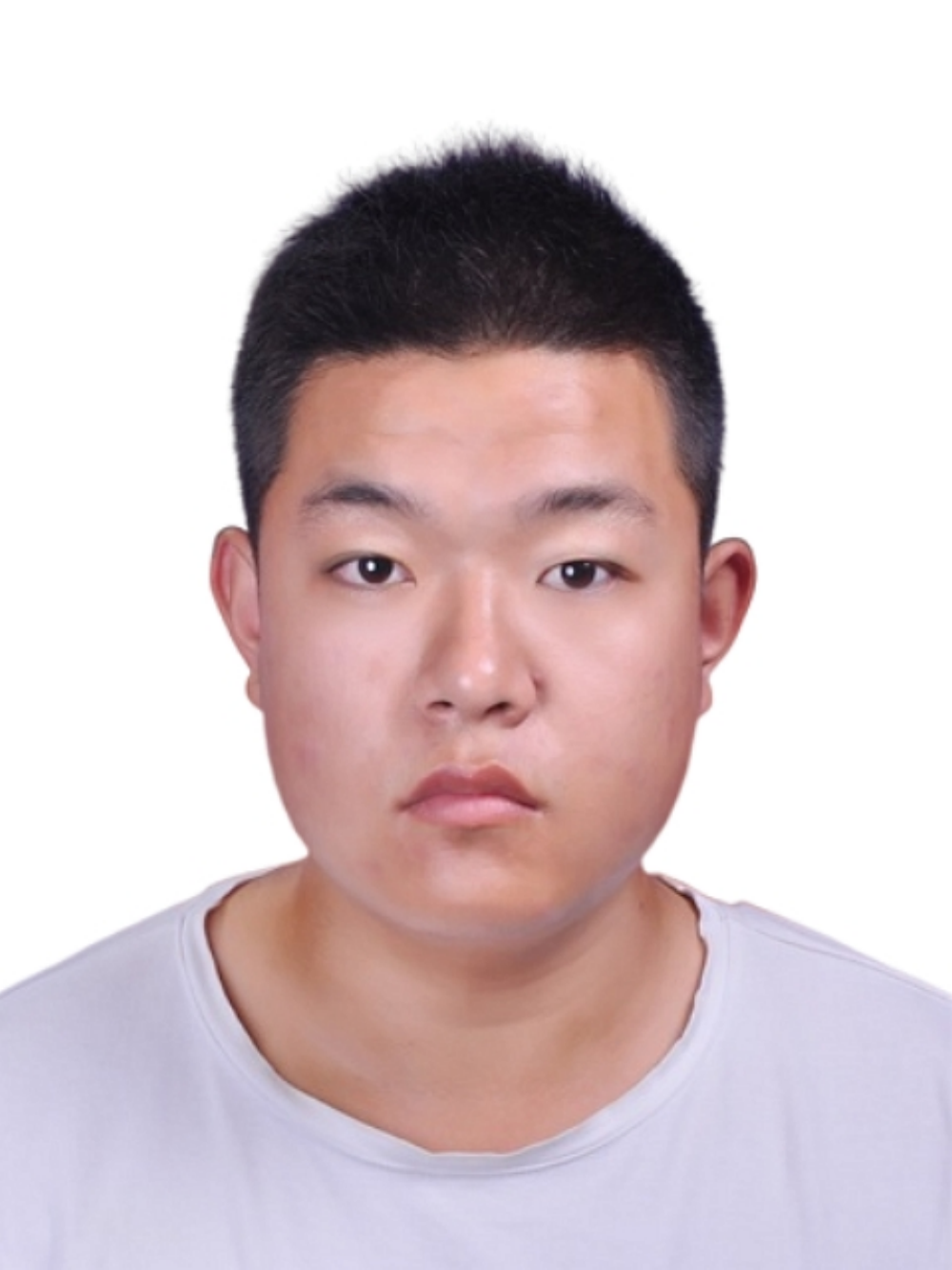}}]{Qiyan Li}
received his Bachelor's degree in electronic information engineering from the University of Electronic Science and Technology of China, Chengdu, China, in 2023. He is currently pursuing a master's degree in electronic information at the University of Electronic Science and Technology of China. His research interests include lidar SLAM and multi-sensor fusion localization and navigation.
\end{IEEEbiography}

\end{document}